\documentclass[article]{beg_32}             %  final version
\usepackage[hang]{footmisc}
\usepackage{hyperref}
\usepackage{amsmath}
\renewcommand{\dmyy}{20}
\renewcommand{\myyear}{2024}
\renewcommand{\today}{}

\usepackage{mathtools}
\usepackage[normalem]{ulem}

\def\R{{\mathbb R}}

\def\d{{\mathbf d}}
\def\q{{\mathbf q}}
\def\r{{\mathbf r}}
\def\w{{\mathbf w}}

\def\s{{\mathbf s}}
\usepackage{tikz-cd}
\def\jname{{}}

\usepackage[geometry]{ifsym}

\begin{document}

\volume{}
\title{Stratospheric aerosol source inversion: Noise, variability, and uncertainty quantification}
\titlehead{Stratospheric aerosol source inversion}
\authorhead{J. Hart, I. Manickam, M. Gulian , L. Swiler, D. Bull, T. Ehrmann, H. Brown, B. Wagman, \& J. Watkins}
%For at least  authors with different addresses, use instead the following commands
\corrauthor[1]{J. Hart}
\author[1]{I. Manickam}
\author[2]{M. Gulian}
\author[1]{L. Swiler}
\author[1]{D. Bull}
\author[1]{T. Ehrmann}
\author[1, 3]{H. Brown}
\author[1]{B. Wagman}
\author[2]{J. Watkins}
\corremail{joshart@sandia.gov}
\corraddress{Sandia National Laboratories, P.O. Box 5800, Albuquerque, NM 87123-1320}
\address[1]{Sandia National Laboratories, P.O. Box 5800, Albuquerque, NM 87123-1320}
\address[2]{Sandia National Laboratories, PO Box 969, Livermore, CA 94551-0969}
\address[3]{Department of Atmospheric Science, University of Wyoming, Laramie, WY, 82071}
% End information for at least  authors with different addresses
% For authors with the same post address,
%\corrauthor{First A. Author}
%\corremail{f.author@affiliation.com}
%\author{Second B. Author, Jr.}
%\address{Department of Chemistry and Courant, Institute of Mathematical Sciences, New York, NY 10012, USA}
% End commands for all authors with the same address

%\dataO{mm/dd/yyyy}
%\dataO{}
%\dataF{mm/dd/yyyy}
%\dataF{}

\abstract{
Stratospheric aerosols play an important role in the earth system and can affect the climate on timescales of months to years. However, estimating the characteristics of partially observed aerosol injections, such as those from volcanic eruptions, is fraught with uncertainties. This article presents a framework for stratospheric aerosol source inversion which accounts for background aerosol noise and earth system internal variability via a Bayesian approximation error approach. We leverage specially designed earth system model simulations using the Energy Exascale Earth System Model (E3SM). A comprehensive framework for data generation, data processing, dimension reduction, operator learning, and Bayesian inversion is presented where each component of the framework is designed to address particular challenges in stratospheric modeling on the global scale. We present numerical results using synthesized observational data to rigorously assess the ability of our approach to estimate aerosol sources and associate uncertainty with those estimates.
}


\maketitle

\section{Introduction}

Stratospheric aerosols from volcanic eruptions can significantly alter regional or global climate patterns on the time scale of months or even years. The most notable eruption with modern observational data, that of Mount Pinatubo in 1991, resulted in a warming of the lower stratosphere by more than 2.5 degrees $C^\circ$ \citep{stenchikov1998,kremser2016,timmreck2012,robock2000,ramachandran2000}. The extent of the climate impact is proportional to the eruption magnitude \citep{marshall2019}; however, many confounding processes in the atmosphere make it difficult to attribute precisely how much a climate anomaly is due to the volcanic aerosols.  Downstream climatic impacts tend to be well observed, but are intermixed with multiple other forcings in the system (e.g. anthropogenic emissions, natural and internal variability, local factors, etc.). Furthermore, full characterization of the eruption itself is challenging as direct observations of the aerosols are not always readily available. This motivates the use of inverse uncertainty quantification (UQ) methods to estimate the eruption characteristics. We represent the eruption probabilistically via a Bayesian formulation. Samples from the posterior distribution enable forward UQ to support rigorous analysis of the extent to which a particular climate impact can be attributed to the volcanic eruption rather than variability within the earth system. In this article, we present a mathematical framework enabling Bayesian inversion to estimate the Mt. Pinatubo volcano source characteristics. Our approach has broader potential application; however, we focus on the problem of volcanic aerosols to motivate the problem characteristics which shape our proposed approach.  

The inverse problem under consideration is posed as source estimation for a system of partial differential equations modeling the earth system. The forward model used for this problem, which models aerosols evolving in time, is the Energy Exascale Earth System Model (E3SM) \citep{golaz2022}, modified to enable prognostic stratospheric aerosols (E3SMv2-SPA) \citep{hbrown2024}. The computational complexity of E3SM inhibits directly using it for source inversion, and necessitates the development of a surrogate to enable the inversion. Using surrogates to enable calibration of earth system models is common in practice \citep{ray2015,li2018,ricciuto2018,yarger2024}. However, in most cases the surrogate is fit to low-dimensional quantities of interest corresponding to spatial and temporal averages of the model prediction. In our case, finer spatial and temporal resolution is required to exploit the resolution of the satellite data and time scales characteristic of the aerosol transport. We will leverage recent advances in operator learning to construct operator neural networks tailored to the structure of our problem. Our focus in this article is the interplay between the operator learning approach and its use to enable Bayesian inversion. Particular emphasis is given to addressing the challenge of atmospheric variability stemming from the chaotic flow characteristics of the wind and imprecision of wind data in the stratosphere.

Traditional surrogate modeling seeks to approximate a function mapping between finite dimensional spaces (frequently Euclidean spaces), but is typically limited by the curse of dimensionality. Operator learning is a burgeoning field which seeks to approximate an operator mapping between infinite dimensional function spaces. From this perspective, operator learning may be viewed as surrogate modeling in the limit as the dimension goes to infinity. Although the curse of dimensionality cannot be avoided, its severity can be lessened by exploiting the mathematical structure of function spaces to design approximations which scale more effectively. Many operator learning approaches may be viewed as a combination of dimension reduction on function spaces and regression mapping between the reduced dimensions. The key idea is that tailoring the dimension reduction and regression to known characteristics of the operator can enable efficient learning. Many operator learning methods are based on deep neural network (DNN) approximations. Examples include modal methods \citep{patel2018nonlinear,patel2021physics,qin2019data,qin2021data,li2020neural,fno_2020}, graph based methods \citep{trask2022enforcing, gao2020physics, shukla2022scalable}, principal component analysis based methods \citep{bhattacharya2020model, hesthaven_2018}, meshless methods \citep{trask2019gmlsnets}, trunk-branch based methods \citep{lu2019deeponet,cai2021deepm}, and time-stepping methods \citep{You2021,Long2018,qin2019data,qin2021data}. Other approaches include manifold learning using Polynomial Chaos models~\citep{Kontolati_2022,KONTOLATI2022111313},  Gaussian Process models~\citep{GIOVANIS2020113269}, and models based on polynomial approximations of reduced operators~\citep{opinf_peherstorfer_2016,opinf_mcquarrie_2021}. The best method for a given problem is based on the function spaces under consideration, characteristics of the operator being approximated, and the amount of data available for training. Since the number of climate simulations available for training is limited, we develop an operator learning approach tailored to characteristics of stratospheric aerosol transport. Specifically, we design a spatial dimension reduction approach to efficiently capture aerosol plume advection and use a model architecture that enforces physical constraints derived from chemistry. 

Inverse problems arise in many areas across the geosciences. Examples include estimation of basal dynamics of the Antarctic ice sheet~\citep{isaac2015scalable}, identification of contaminant sources in the subsurface~\citep{laird_2005}, modeling of plate mechanics in mantle flow~\citep{ratnaswamy_2015}, reconstruction of subsurface material properties via full waveform inversion~\citep{fwi_2014}, and inference of sources in atmospheric transport~\citep{enting2002inverse}. Due to the high-dimensionality created by spatial heterogeneity, it is common to use optimization and derivative-based sampling algorithms to achieve computationally scalable algorithms~\citep{ghattas_infinite_dim_bayes_1,ghattas_infinite_dim_bayes_2,cui_2014,Biegler_11}. Considerable research has gone into atmospheric source inversion with a focus on greenhouse gas emissions~\citep{ray_2014,deng_2022}. This article focuses on a different class of atmospheric source inversion: stratospheric aerosol inversion, which is characterized by a point source injection advected globally by stratospheric winds.

Our contributions in this article are:
\begin{enumerate}
\item[$\bullet$] a novel operator learning approach that models aerosol transport from data with varying atmospheric states and injection masses using nonlinear spatial dimension reduction via radial basis functions and a chemistry-informed architecture,
\item[$\bullet$] a novel Bayesian approximation error approach to accommodate both internal atmospheric variability and background aerosols in volcano source inversion,
\item[$\bullet$] an application informed framework which couples earth system simulation, operator learning, and inverse problems,
\item[$\bullet$] a demonstration of our proposed framework using hold-out simulation data from unseen injection mass and atmospheric states as synthesized satellite observations to rigorously test the approach. 
\end{enumerate}
Limitations on data generation due to the computational complexity of earth system models is a central challenge we consider. Our approach is tailored to ensure feasibility in such limited data scenarios. This article does not seek to solve the inverse problem using observational aerosol data, but rather we stress test our proposed approach using unseen simulation data as observations. Such testing is a crucial prerequisite to using satellite data.

\section{Overview of the problem}

Volcanic aerosol evolution is a well-researched process that begins with the injection of $SO_2$ gas into the stratosphere, see~\citep{Zanchettin2022,timmreck2012,robock2000} for comprehensive overviews. Through chemical processes, this gas is transformed into sulfate aerosol which grows larger through microphysical processes and reflects/absorbs solar radiation causing changes in stratospheric and surface temperatures. Satellite data provides a measure of how much sunlight is scattered and absorbed by the aerosols and is quantified by aerosol optical depth (AOD), which is an aggregate quantity from all aerosols present within a column of the atmosphere observed by the satellite \footnote{Alternatives to AOD observations include instruments for species-specific characterization. However, AOD is a robust and generic indicator of aerosol change. Restricting ourselves to AOD measurements ensures generality and extensibility of our approach.}.

Let $\Omega \subset \R^3$ denote the spatial domain of the earth system model which is defined by coordinates of longitude, latitude, and altitude\footnote{E3SM uses pressure level rather than altitude as a coordinate, but we use the term altitude for simplicity of the exposition.}, and let $[0,T]$ denote the time interval under consideration (which is on the order of days to weeks for this problem). Let $u:\Omega \times [0,T] \to \R^m$ denote the vector-valued function of state variables in the earth system model, which can be expressed as
\begin{align} \label{eqn:forward_model}
& \dot{u} = f(u) + g e_{SO_2}\\
& u(0) = u_0, \nonumber
\end{align}
where $f(u):\Omega \times [0,T] \to \R^m$ models the dynamics, $g:\Omega \times [0,T] \to \R$ models the $SO_2$ injection, $e_{SO_2} \in \R^m$ is a vector with $1$ in the entry corresponding to the $SO_2$ state and $0$ otherwise, and $u_0:\Omega \to \R^m$ is the initial state.

There are many variables in the coupled earth system, i.e. $m$ is large, due to the many coupled processes in the earth system. For our problem, the most important variables are:
\begin{enumerate}
\item[$\bullet$] the mass of $SO_2$, 
\item[$\bullet$] the mass of sulfate aerosol, 
\item[$\bullet$] the aerosol optimal depth (AOD), and
\item[$\bullet$] the zonal wind (wind in the direction parallel the equator).
\end{enumerate}

\textbf{Our goal is to:}
\begin{enumerate}
\item[(i)] learn a surrogate model which takes the zonal wind and $SO_2$ source as inputs, and predicts the evolution from $SO_2$ to sulfate to AOD,
\item[(ii)] use this surrogate to constrain an inverse problem that infers the $SO_2$ source from AOD observations.
\end{enumerate}

Observational data collected via satellites is at a fine spatio-temporal resolution ($\mathcal O(1)$ degrees longitude and $\mathcal O(1)$ days). However, many atmospheric analyses are done at a coarser resolution to reduce complexity. Such coarsening simplifies the modeling but results in loss of information to inform the inverse problem. This trade-off is a frequent challenge in inverse problems as making the forward problem easier (via smoothing) makes the inverse problem more challenging (more ill-posed). We seek to formulate the inverse problem on the time scale of weeks with moderate spatial averaging/smoothing to preserve the richness of information content in the fine resolution satellite data. 

In addition to this classical model complexity versus information content trade-off, many challenges arise from unique characteristics of this volcanic aerosol inverse problem. These include:
\begin{enumerate}
\item[$\bullet$] variability in the atmosphere, and hence the zonal wind, due to initial state uncertainty and coupled processes in the earth system,
\item[$\bullet$] local-to-global spatial scales as the volcanic eruption is initially localized in space but spreads equatorially around the globe in approximately 3 weeks, and
\item[$\bullet$] the presence of background aerosols which contribute to the AOD but do not come from the volcano. 
\end{enumerate}

\begin{figure}[h]
\centering
  \includegraphics[width=0.99\textwidth]{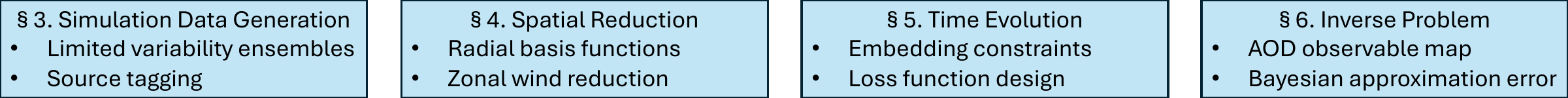}
    \caption{Overview of the article's organization. Each bullet point highlights an important aspect of the proposed framework.}
  \label{fig:overview_figure}
\end{figure}

We propose a novel combination of techniques, which in themselves are well established, but whose adaptation and composition is guided by the challenges outlined above. Figure~\ref{fig:overview_figure} provides an overview of the various aspects of our approach and how they are organized in the article. In the sections which follow we detail each aspect and highlight how it relates to the larger framework. Specifically, in Section~\ref{sec:simulation} we consider the nuances of the earth system simulations needed to facilitate our analysis. Section~\ref{sec:spatial_reduction} presents our approach to spatial dimension reduction, which prepares reduced data as inputs for learning a time evolution operator in Section~\ref{sec:time_evol}. The combination of spatial encoding and time evolution of the reduced spatial coordinates defines a reduced order model for the aerosol transport. In Section~\ref{sec:bayes_inv} we formulate a Bayesian inverse problem using the reduced model and present a Bayesian approximation error approach to incorporate uncertainty from background aerosols and atmospheric variability in the inverse problem. Numerical results are given in Section~\ref{sec:numerical_results} to demonstrate our approach using data from E3SM. Concluding remarks are made in Section~\ref{sec:conclusion}.

\section{Simulation data generation and processing} \label{sec:simulation}

To generate training data, the forward model~\eqref{eqn:forward_model} will be solved for various sources $g$ and initial states $u_0$. We are limited to $\mathcal O(30)$ forward model evaluations, so our design of source and initial states for these simulations is crucial to enable reliable reduced order models within the range of variability needed for the inversion. In the subsections which follow we detail our use of limited variability ensembles (the design of $u_0$), source tagging (tracking of aerosols from $g$), and extraction of the relevant states to form the training set. We conclude this section with an overview of our inverse problem formulation and algorithmic framework.

\subsection{Limited variability ensembles}
Traditional earth system modeling accounts for initial state uncertainty by running \emph{ensembles} of model evaluations, that is, evaluating the model for different initial states~\citep{esd-14-1107-2023}. Typically, these initial states are designed to be statistically independent so that the set of ensemble members captures the full range of possible system states. We refer to this ensemble design as full variability. However, in our context, we analyze a volcanic eruption which occurred in the past and some knowledge of the atmospheric state is available. This motivates the use of limited variability ensembles which constrain the initial states $u_0$ to be representative of the atmosphere as it was partially observed at the time of the volcanic eruption. 

To design a limited variability ensemble, five full variability ensemble members are generated by randomly perturbing (with values near machine precision) the temperature field of a historical simulation starting in 1985. Running simulations until June 1991, right before the Mt. Pinatubo eruption, yields five ensemble members whose June 1991 states are statistically independent. We select a ``best" ensemble member that most closely matches observed climate modes (as determined from reanalysis products). Specifically, we match the 1991 El Ni\~no and the Quasi Biennial Oscillation (QBO) modes. Four metrics are used to determine the best match: the NINO3.4 value and  NINO3.4 trend, and the QBO phase at the atmospheric pressure levels of 10 and 50 hectopascal \citep{ehrmann2024}. The limited variability ensemble members are generated by perturbing the initial temperature field (with near-machine precision values) of the best-matched full variability ensemble member. Since this perturbation occurs shortly before the volcanic eruption, all of the limited variability ensemble members match the 1991 climate modes. Yet, these near machine precision perturbations still induce nontrivial variability in the stratospheric winds. Accounting for this variability is a key aspect of our approach to stratospheric aerosol inversion.

\subsection{Source tagging}
The observable variable AOD has contributions from diverse sources including volcanos, dust storms, industrial processes, etc. The background aerosols (all aerosols except for the volcanic aerosols) create additional challenges for volcanic aerosol source inversion since the rapid transport and mixing of aerosols makes it difficult to disentangle the various sources. We leverage an aerosol source tagging method~\citep{yang_2024} within E3SM which provides the capability to separate aerosol tracers by emission source and evolves them separately in the forward model. Specifically, for a state variable $\zeta$ modeling a chemical species, source tagging represents $\zeta = \zeta_v + \zeta_b$, where $\zeta_v$ is the species due to the volcanic source and $\zeta_b$ is the background species due to all other sources. Separate differential equations evolve these two species, thus ``tagging" which aerosols come from the volcano. This decomposition into volcanic and background species is done for the $SO_2$, sulfate, and AOD variables.

\subsection{Summary of simulation data}
The forward model~\eqref{eqn:forward_model} is evaluated for $N_e$ ensembles (perturbed initial states $u_0$) and $N_s$ sources $g$. The sources are spatially localized around the volcano by defining the spatial profile of $g$ as a Dirac delta function. The temporal profile of $g$ is defined over a $9$ hour time window during the eruption, after which time $g=0$. The $N_s$ sources differ by their injection magnitudes which are chosen to capture a realistic range of plausible volcanic eruptions. Due to the short time-scale of the eruption relative to the time-scale of our analysis (hours compared to days), we pose the inverse problem to estimate the source tagged $SO_2$ shortly after the eruption ends. We restrict our analysis to the most relevant state variables and consider the dataset 
\begin{align}
\label{eqn:raw_data}
\{ \alpha_v^{i,j},\beta_v^{i,j},\rho_v^{i,j},\rho_b^{i,j},\omega^{i,j} \}_{i=1,j=1}^{N_e,N_s} 
\end{align}
which corresponds to the $SO_2$ ($\alpha$), sulfate ($\beta$), AOD ($\rho$), and zonal wind ($\omega$). Each state is indexed by $i,j$ to identify which ensemble it arises from. The subscript $v$ indicates the volcanic source tags and $\rho_b$ is the background AOD variable that will be used to incorporate background aerosol data in the inverse problem. The forward model output is on a daily time scale and hence our data is evaluated at time steps $t_k \in \{0,1,2,\dots,N_t\}$ on a horizon of $N_t+1$ days where $t_0$ is 24 hours after the onset of the eruption. The $SO_2$, sulfate, and zonal wind variables are functions defined in three spatial dimensions, while the AOD variables are defined in two spatial dimensions, since AOD is a column integrated quantity. 

\begin{figure}[h]
\centering
  \includegraphics[width=0.8\textwidth]{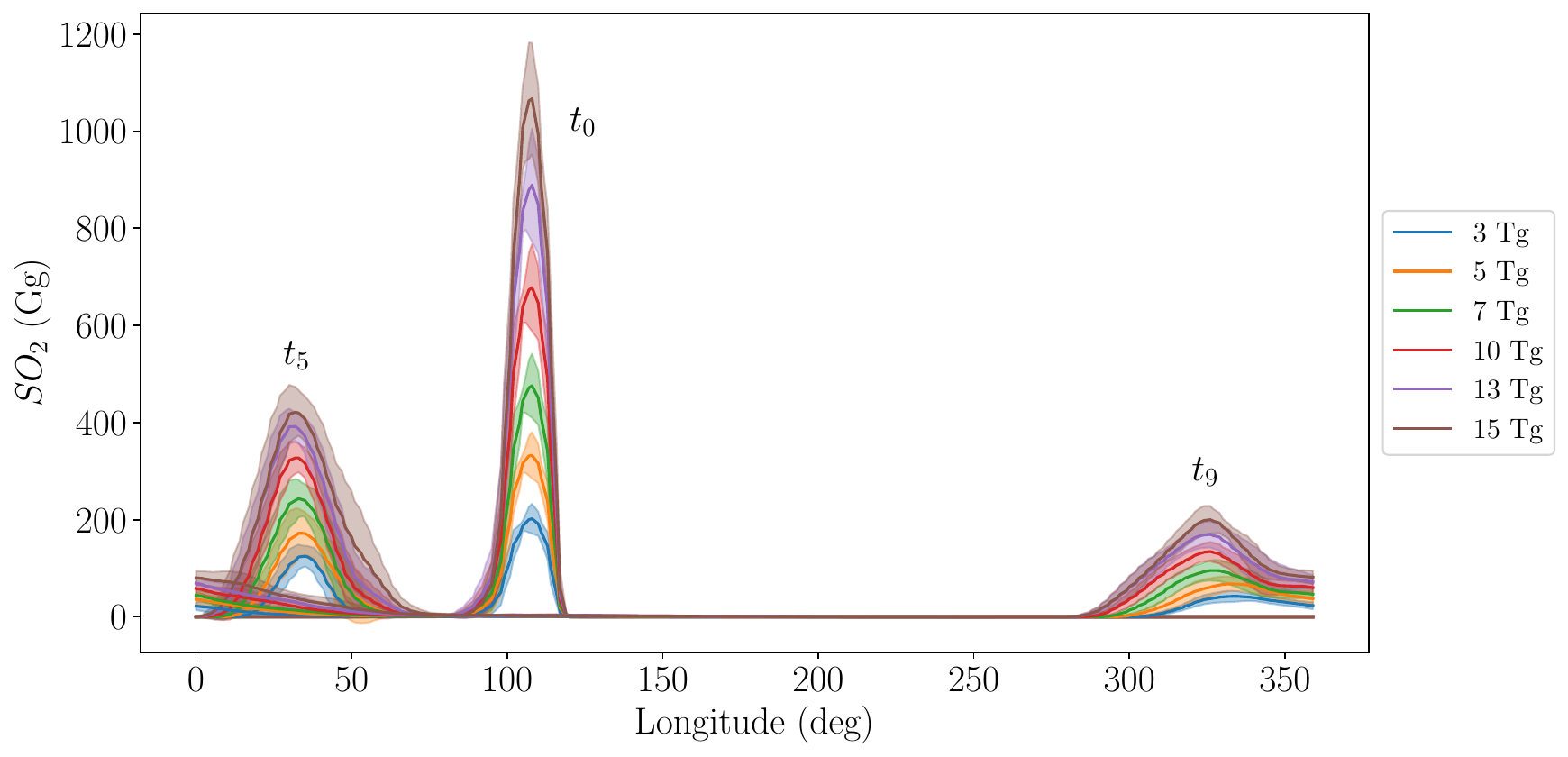}
    \caption{Source tagged 1D profile of $SO_2$, as a function of longitude, at time steps $t_0$, $t_5$, and $t_9$. The color indicates the volcano injection mass. At each time step, the solid lines correspond to the ensemble mean of the $SO_2$ and the shading indicates two standard deviations.}
  \label{fig:so2_ensembles}
\end{figure}

Figure~\ref{fig:so2_ensembles} displays source tagged $SO_2$ to illustrate the characteristics of its transport and variability. In particular, Figure~\ref{fig:so2_ensembles} shows three different time steps. At each time step, we compute the mean and standard deviation over ensembles to demonstrate how the ensemble variability compares with eruption magnitude variability.

\subsection{Overview of the proposed inversion framework}

\begin{figure}[h]
\centering
  \includegraphics[width=0.95\textwidth]{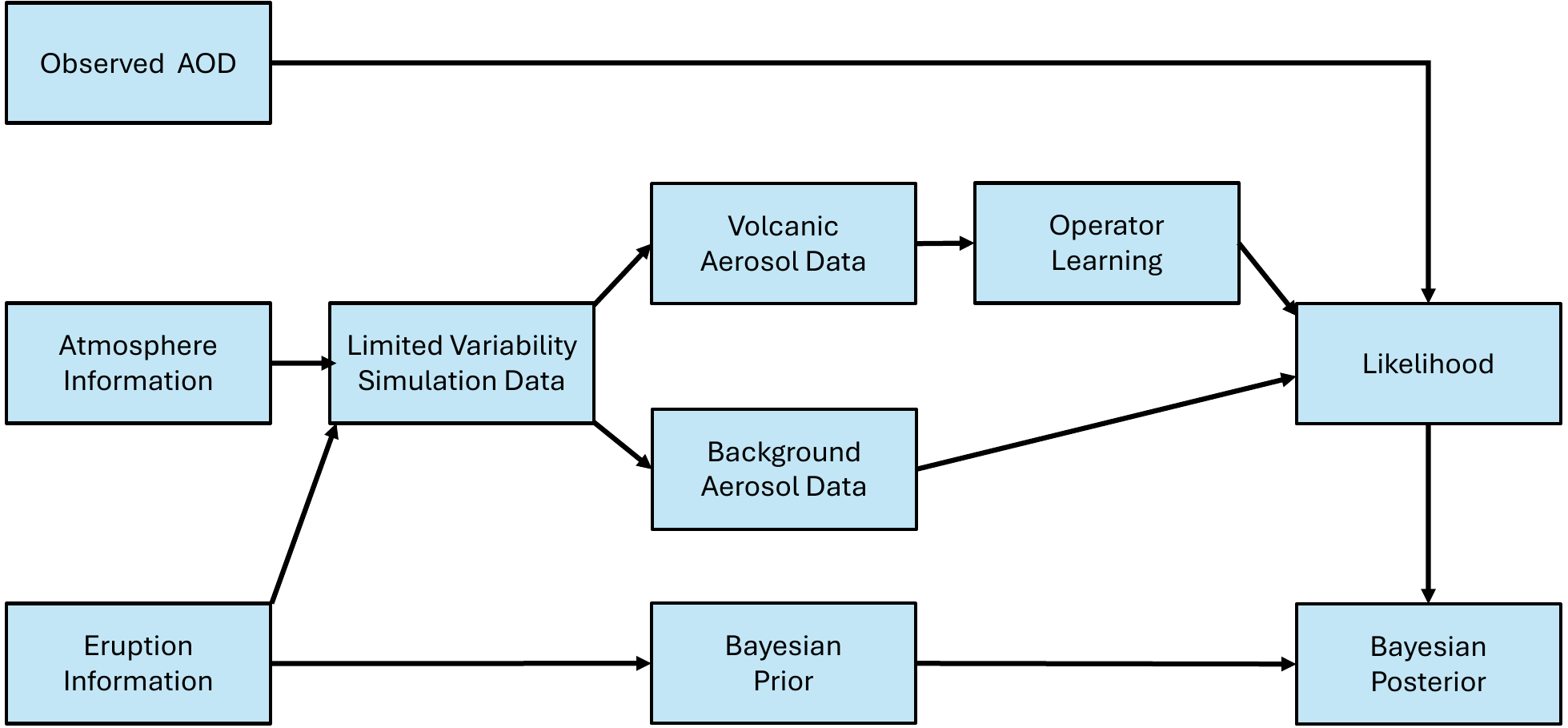}
    \caption{Overview of the proposed framework.}
  \label{fig:algorithm_overview_figure}
\end{figure}

We formulate an inverse problem to estimate the volcanic source tagged $SO_2$ (i.e. $\alpha_v$) at the initial time $t_0$, using observations of AOD without source tags (i.e. $\rho_v+\rho_b$) at later times $t_k$, $k=1,2,\dots,N_t$. Note that this is an initial condition inversion rather than forcing term inversion because of the time scales in the data. That is, we are estimating the $SO_2$ component of the initial state, where $t_0$ corresponds to the time shortly after the eruption has ended. We use operator learning to construct a reduced model to evolve $\alpha_v$ to $\beta_v$ in time and a separate stationary model to map $\beta_v$ to $\rho_v$. The background AOD $\rho_b$ is incorporated into the inverse problem in our formulation of the likelihood function. Figure~\ref{fig:algorithm_overview_figure} overviews the important components of our proposed framework and how they interact with one another.

\section{Spatial dimension reduction} \label{sec:spatial_reduction}

\begin{figure}[h]
\centering
\includegraphics[width=0.99\textwidth]{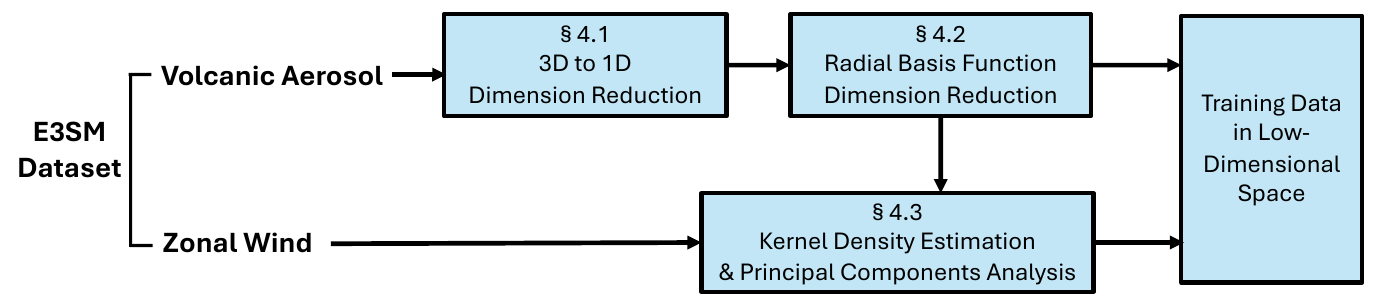}
    \caption{Overview of the spatial dimension reduction approach for aerosol and wind data.}
  \label{fig:preprocess_pipeline}
\end{figure}

The raw data~\eqref{eqn:raw_data} is large due to its four dimensions (three spatial dimensions and time). In the small data setting, i.e. $N_eN_s =\mathcal O(30)$, it is intractable to train an operator surrogate for complex and high dimensional dynamics. Rather, we seek to reduce the spatial dimension and train a time evolution operator in the low-dimensional space. Spatial dimension reduction is multifaceted as we consider both data preprocessing, aerosol dimension reduction, and wind dimension reduction. Figure~\ref{fig:preprocess_pipeline} shows an overview of the preprocessing steps performed on the inputs to the time evolution operator. The subsections which follow consider these three facets. 

\subsection{Data preprocessing}
\begin{figure}[h]
\centering
  \includegraphics[width=0.99\textwidth]{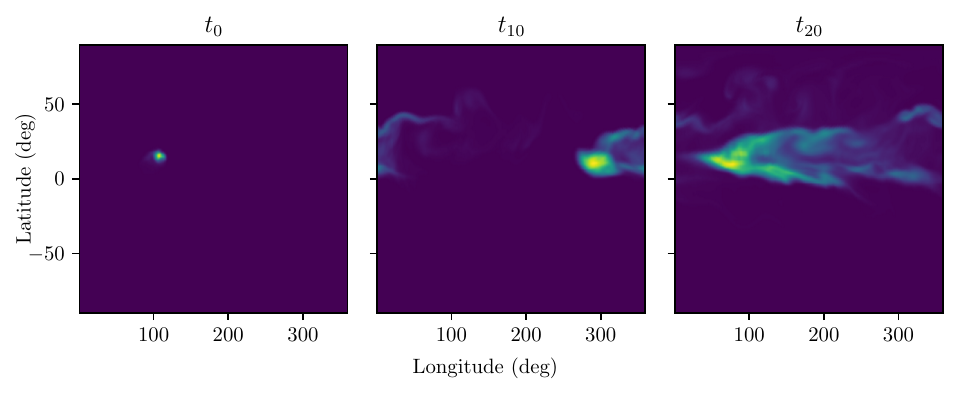}
    \caption{Representative simulation of the aerosol transport in two dimensions. The panels from left to right correspond to time steps at $t_0$, $t_{10}$, and $t_{20}$, i.e. a $21$ day period. Longitude is measured eastward from the Greenwich prime meridian.}
  \label{fig:sulfate_2d}
\end{figure}

Although the data is inherently three dimensional in space, the dynamics are faster in the longitudinal directions as prevailing winds drive the aerosols around the globe equatorially in the first three weeks post-eruption. This indicates that integrating over the latitude and altitude directions can significantly reduce the dimension while preserving important dynamical characteristics. In general, integrating out spatial dimensions has the effect that learning a surrogate for the forward problem becomes easier, but the information lost in the integration makes the inverse problem more challenging. This trade-off of dynamical complexity and information content is crucial and it highlights the need to customize the forward model surrogate with a cognizance of the inverse problem. Our results indicate that compression to only longitude dependence is effective for volcanic aerosol transport because the longitudinal transport is much faster compared to the other dimensions. Figure~\ref{fig:sulfate_2d} shows a representative simulation of the aerosol transport in two dimensions (latitude and longitude). The aerosols are transported equatorially around the globe in approximately 21 days ($t_0$ to $t_{20}$) while remaining contained within a latitudinal band around the tropics. The aerosols eventually reach high latitudes after multiple months. However, the inverse problem is best informed by the transport in the early days post-eruption.

\subsection{Radial basis function dimension reduction}
Let $\xi:\Omega_\text{lon} \times [0,T] \to \R$, $\Omega_\text{lon} \subset \R$, denote an arbitrary volcanic source tagged aerosol variable ($SO_2$, sulfate, or AOD) after integrating over the latitude and altitude dimensions. We seek to further reduce the spatial dimension of $\xi$ by encoding it in a low-dimensional spatial basis so that we can train a time evolution operator in the reduced dimension. All aerosol variables under consideration have the spatio-temporal characteristic that they are spatially localized at the initial time and are transported around the globe by the advective force of the zonal winds, as seen in Figure~\ref{fig:so2_ensembles}. Such advection dominated dynamics is known to be challenging for linear dimension reduction methods (such as principle component analysis) and hence we consider nonlinear dimension reduction. There are a multitude of potential approaches for nonlinear encoding of spatial fields~\citep{KONTOLATI2022111313,LEE2020108973}.

In this work we consider Gaussian radial basis functions (RBFs). This choice is motivated by two characteristics of the volcanic aerosol plume: (i) the plume has an approximate Gaussian (bell curve) shape, and (ii) the plume exhibits non-smooth, small amplitude features which are smoothed out by the RBFs, thus eliminating spatial scales which cannot be learned from limited data. We consider spatial basis functions of the form
\begin{align}
\label{eqn:gaussian_basis}
\psi_\ell(x) = c_\ell \exp\left( -a_\ell^2 \mid x - x_\ell \mid^2 \right)
\end{align}
where $x_\ell, a_\ell,c_\ell \in \R$ are the center, shape, and coefficient hyperparameters respectively, which will be fit via nonlinear least squares. 

The aerosol variable $\xi(x,t)$ is periodic in $x$ (since $x$ is the longitude coordinate) and hence we must periodize~\eqref{eqn:gaussian_basis}. Following~\citep{xiao_thesis}, we consider basis functions
\begin{align}
\label{eqn:gaussian_basis_periodic}
\Psi_\ell(x) = \sum\limits_{m=-\infty}^\infty \psi_\ell(x+mL)
\end{align}
where $L$ is the period (if $x$ is longitude measured in degrees then $x \in \Omega_\text{lon}=[0,360]$ and $L=360$). In practice, the infinite sum in~\eqref{eqn:gaussian_basis_periodic} can be truncated to $m=-M,\dots,M$, where $M$ is chosen based on the shape hyperparameter $a_\ell$. Truncation errors are generally on the order of machine precision and hence are negligible. 

Given $\xi(x,t_k)$, for an arbitrary time step $t_k$, we consider a $N_{rbf}$ dimensional basis $\{ \Psi_\ell \}_{\ell=1}^{N_{rbf}}$ and fit the center, shape, and coefficient hyperparameters. That is, we approximate 
\begin{align}
\label{eqn:rbf_fit}
\xi(x,t_k) \approx \sum\limits_{\ell=1}^{N_{rbf}} \Psi_\ell(x;x_\ell^k,a_\ell^k,c_\ell^k)
\end{align}
where $N_{rbf}$ is the number of basis functions and the hyperparameters are indexed with a superscript $k$ to indicate the time step. This gives a $3 N_{rbf}$ dimensional embedding of $\xi(x,t_k)$. This is done for each time step, resulting time series $\{ x_\ell^k, a_\ell^k,c_\ell^k \}_{k=0}^{N_t}$, $\ell=1,2,\dots,N_{rbf}$, which will be used as training data to learn a time evolution operator in the space of the RBF hyperparameters. 

Traditional use of RBFs considers a fixed grid of centers $\{x_\ell\}_{\ell=1}^{N_{rbf}}$, chooses optimal shape hyperparameters $\{a_\ell \}_{\ell=1}^{N_{rbf}}$ based on the placement of the centers, and then fits the coefficients $\{ c_\ell \}_{\ell=1}^{N_{rbf}}$ via linear least squares. Our approach differs: rather than taking $N_{rbf}$ large enough for $\{x_\ell\}_{\ell=1}^{N_{rbf}}$ to cover $\Omega_\text{lon}$, we instead take a small $N_{rbf}$ and fit the coefficient, center, and shape hyperparameters. This gives a nonlinear embedding which is able to capture advective phenomena in a low-dimensional space by permitting the center hyperparameter to evolve in time. Our approach introduces a challenge of identifiability as multiple sets of hyperparameters may yield identical or nearly identical approximations. However, this can be addressed by judiciously selecting a small $N_{rbf}$ and enforcing constraints on the hyperparameters.

\subsection{Wind dimension reduction} \label{ssec:wind_dim_reduction}

To incorporate zonal wind into the aerosol plume time evolution operator, we require a low-dimensional embedding of $\omega$ that is localized about the aerosol plume. However, an RBF approximation of $\omega$ is not suitable since the zonal wind is not spatially localized like the sourced tagged aerosols.

The core idea that makes our RBF approach effective is that the center hyperparameter is permitted to move and hence the basis function retains low dimensionality as it is advected around the globe. To impart spatial locality to the zonal wind, we weight the wind data using the aerosol mass and thus restrict it to a local region in the atmosphere. Specifically, a threshold $\tau_{SO_2}$ is specified and used to define a time varying set 
\begin{align}
\label{eqn:wind_tau}
\mathcal D(t) = \{(x,y,z) \in \Omega \mid \alpha_v(x,y,z,t) \ge \tau_{SO_2} \}
\end{align}
which restricts the spatial domain $\Omega \subset \R^3$ to the region where the volcanic aerosol plume has a magnitude greater than the threshold $ \tau_{SO_2}$. Note that $(x,y,z)$ corresponds to longitude, latitude, and altitude. For each time step, we introduce RBF-based weighting functions
\begin{align}
\label{eqn:wind_rbf}
\phi_\ell^k(x,y,z) = \Psi_\ell(x;x_\ell^k,a_\ell^k,c_\ell^k) \chi_{\mathcal D(t_k)}(x,y,z)
\end{align}
$\ell=1,2,\dots,N_{rbf}$, where $\chi$ is the indicator function of a set. Hence $\phi_\ell^k \ge 0$ captures the spatial locality of the RBF basis functions by weighting locations in space by the amount of aerosol being represented in the $\ell^{th}$ RBF basis function at the $k^{th}$ time step. 

For each time step $t_k$, we consider point-wise spatial evaluations of the zonal wind\\ $\omega(x,y,z,t_k)$ as samples of a random variable. Weighting these samples with $\phi_\ell^k(x,y,z)$ defines a distribution of zonal wind values that is localized about the RBF basis function $\Psi_\ell$ at time $t_k$. Using weighted kernel density estimation, we produce PDFs $h_\ell^k:\R \to \R$ for the set of zonal winds which confers the highest probability to zonal winds that are characteristic of the region where the aerosol plume is concentrated.

Transforming $\omega$ to $h_\ell^k$ does not reduce the dimension, but rather changes domains as $\omega$ is a function of space and $h_\ell^k$ is a PDF of zonal wind values. This achieves localization commensurate to what is done via our RBF approximation. However, $h_\ell^k$ is still a high dimensional representation of the zonal wind. To compress it, we use principle component analysis~\citep{brunton2022data} to project the PDFs $\{ h_\ell^k \}_{\ell=1,k=1}^{N_{rbf},N_t}$ onto a low-dimensional subspace. Specifically, we learn a collection of principle components $\eta_i:\R \to \R$, $i=1,2,\dots$ and express a particular PDF $h_\ell^k:\R \to \R$ in terms of its principle components as
\begin{align*}
h_\ell^k = \sum_{i} w_i^{\ell,k} \eta_i,
\end{align*}
where $w_i^{\ell,k} \in \R$ are the principle component coordinates for $h_\ell^k$. We truncate to the leading $N_w$ principle components and represent $h_\ell^k$ via its coordinates $\w_\ell^k=(w_1^{\ell,k},w_2^{\ell,k},\dots,w_{N_w}^{\ell,k}) \in \R^{N_w}$. 

We can take small values for $N_w$ since our aggregation of time steps with a moving region eliminates advective characteristics in the data. Furthermore, taking a small $N_w$ has the effect of smoothing the wind. Thus for a given RBF basis function and time step, $\w_\ell^k$ is a $N_w$ dimensional representation of the zonal wind field. This reduces the three dimensional zonal wind field, which has $\mathcal O(10^6)$ degrees of freedom, to a $N_w$ dimensional space, where $N_w = \mathcal O(1)$ in many cases.

\section{Time evolution operator} \label{sec:time_evol}

After spatial dimension reduction, our objective is to train a time evolution operator in the space of reduced coordinates to predict the transport of the volcanic aerosols. The operator needs to accurately trace the evolution of aerosols while accounting for variations in both the volcanic injection magnitudes and atmospheric winds. The architecture and training of the time evolution operator needs to be endowed with awareness of time discretization and enforcement of relevant physical constraints such as the conservation of mass and irreversibility of chemical processes. This is particularly important when training the model with a small dataset; incorporation of physical knowledge compensates for the lack of data. This section considers model architecture and loss function design to achieve this ``physics informed" operator. 

Working with source tagged data simplifies the analysis by avoiding the complexity of other external processes adding or removing aerosols, effectively reducing noise in the training data. Hence our focus is on the volcano-induced $SO_2$, sulfate, and AOD, i.e.  $\{ \alpha_v,\beta_v,\rho_v \}$. This section considers training a time evolution operator to predict the evolution of $\{ \alpha_v,\beta_v\}$ given a zonal wind field $\omega$. The AOD does not evolve dynamically, but rather is computed point-wise in space at a given time instance based on the presence of aerosols in the atmosphere at that time and spatial location. Hence, we model the mapping from $\beta_v$ to $\rho_v$ as an observation operator which we consider in Section~\ref{sec:bayes_inv}. 

 Let $\r_k \in \R^{3 N_{rbf}}$ denote the concatenation of the $SO_2$ RBF hyperparameters $\{ x_\ell^k,a_\ell^k,c_\ell^k\}_{\ell=1}^{N_{rbf}}$ at time $t_k$ and let $\s_k  \in \R^{3 N_{rbf}}$ denote the analogous concatenation of the sulfate RBF hyperparameters. We use the notation $\hat{\r}_k$ and $\hat{\s}_k$ to denote the time evolution operator's prediction of these coordinates. That is, $\{ \r_k,\s_k\}_{k=0}^{N_t}$ denotes the RBF coordinates computed from the data $\{ \alpha_v,\beta_v\}$ and $\{ \hat{\r}_k,\hat{\s}_k\}_{k=0}^{N_t}$ denotes the RBF coordinates predicted by the learned operator.
 
 Another benefit of source tagging is that the mass of sulfur is conserved over time in the volcanic aerosol data. We use this fact to model sulfate as a function of the $SO_2$ at each time step. Initially, the volcano injects $SO_2$ which reacts with hydroxyl radicals in the atmosphere to create sulfate. Letting $M_{\alpha}$ and $M_{\beta}$ denote the molar masses of $SO_2$ and sulfate, respectively, we have that
\begin{align}
\label{eqn:conservation}
\int_{\Omega_{\text{lon}}} \left( \frac{\alpha_v(x,t)}{M_{\alpha}} + \frac{\beta_v(x,t)}{M_{\beta}} \right) dx = \int_{\Omega_{\text{lon}}} \left( \frac{\alpha_v(x,t_0)}{M_{\alpha}} + \frac{\beta_v(x,t_0)}{M_{\beta}} \right) dx
\end{align}
for all $t$. 

The sulfate plume mirrors the spatial profile of the $SO_2$ plume but differs in total mass, so given $SO_2$ at a particular time step, we use~\eqref{eqn:conservation} to predict the sulfate. Specifically, for a fixed time step, we set the center and shape hyperparameters of the sulfate RBF representation to be equal to the $SO_2$ RBF center and shape hyperparameters. The sulfate coefficient hyperparameters are computed by enforcing conservation of mass~\eqref{eqn:conservation}.

Estimation of sulfate via~\eqref{eqn:conservation} requires computing the total mass of sulfur at time  $t_0$. Yet at time $t_0$, which is 15 hours after the volcanic eruption has ended, there is a small amount of sulfate present in the atmosphere due to the reaction of $SO_2$ with hydroxyl radicals. Hence the total mass of sulfur at time $t_0$ is more than just the initial mass of sulfur in the $SO_2$ molecules.  Computing the ratio of sulfate and $SO_2$ masses at $t_0$ in all training set simulations, we observe that the variability of this ratio across the simulations is small due to the brief time window from the eruption to $t_0$. Hence, we compute the mean ratio of masses and use this as a scaling factor\footnote{Across the training set, the ratio of initial sulfate mass to initial $SO_2$ mass range from $0.0533$ to $0.0558$, with a mean of $0.0544$.} to initialize the sulfate mass proportional to the initial $SO_2$ mass. 

Let $\mathcal E$ denote the operator which, for a given time step, computes the sulfate coordinates $\hat{\s}_k$ using the estimated $SO_2$ coordinate $\hat{\r}_k$ and~\eqref{eqn:conservation}, that is, $\hat{\s}_k=\mathcal E(\hat{\r}_k,\hat{\r}_0,\hat{\s}_0)$ for all $k$. This is achieved by enforcing the conservation of molar mass~\eqref{eqn:conservation} for each RBF basis function separately, which translates to conservation of total mass since our approximation is a linear combination of the RBF basis functions. The mass of the RBF basis function can be computed analytically, as shown in~\eqref{eqn:rbf_mass}.

Given that the sulfate can be modeled as a function of the $SO_2$ at each time step, we use a flow map to model the time evolution of the $SO_2$. Let $\w_k \in \R^{N_w N_{rbf}}$ denote the concatenation of the wind coordinates $\{ \w_\ell^k \}_{\ell=1}^{N_{rbf}}$ at time $t_k$. We seek to learn the flow map $\mathcal F:\R^{3 N_{rbf}} \times \R^{N_w N_{rbf}} \to \R^{3 N_{rbf}}$ such that
\begin{align*}
\r_{k+1} \approx \mathcal F(\r_k,\w_k).
\end{align*}
Since the flow map is approximating a differential equation time step, we use an architecture which mimics a forward Euler time stepping scheme,
\begin{align}
\label{eqn:forward_euler}
\mathcal F(\r_k,\w_k) = \r_k + \Delta t \mathcal N(\r_k,\w_k)
\end{align}
where we assume constant time step sizes $\Delta t = t_{k+1}-t_k$ and introduce the model $\mathcal N$ to be learned. Since time and space have been modeled via the forward Euler architecture~\eqref{eqn:forward_euler} and RBF dimension reduction, respectively, we model $\mathcal N:\R^{3 N_{rbf}} \times \R^{N_w N_{rbf}} \to \R^{3 N_{rbf}}$ using a dense feed forward neural network. 

Since no additional $SO_2$ enters the system (due to our use of source tagged data) and $SO_2$ is depleted over time via its reaction with hydroxyl radicals to create sulfate, we require that the mass of $SO_2$ (integrated over the spatial domain) be monotonically decreasing. For a basis function $\Psi_\ell$ as in~\eqref{eqn:gaussian_basis_periodic}, we have 
\begin{align}
\label{eqn:rbf_mass}
\int_{\Omega_\text{lon}} \Psi_\ell(x) dx = \sqrt{\pi}\frac{c_\ell}{a_\ell},
\end{align}
where we assume without loss of generality that $a_\ell>0$. We can enforce monotonicity of the $SO_2$ mass by using a ReLU output layer for $\mathcal N$ that makes the time step increment nonpositive. By embedding this monotonicity structure in the learned operator, we enforce the irreversibility of chemical processes so that $SO_2$ cannot be created over time. Where appropriate, we can also impose monotonicity on the RBF center hyperparameter $x_\ell$ if, for instance, the plume is always flowing from east to west. In general, the forward Euler architecture of~\eqref{eqn:forward_euler} simplifies enforcement of monotonicity constraints  since the constraints are equivalent to non-negativity (or non-positivity) in the output components of $\mathcal N$. 

\begin{figure}[h]
\centering
  \includegraphics[width=0.89\textwidth]{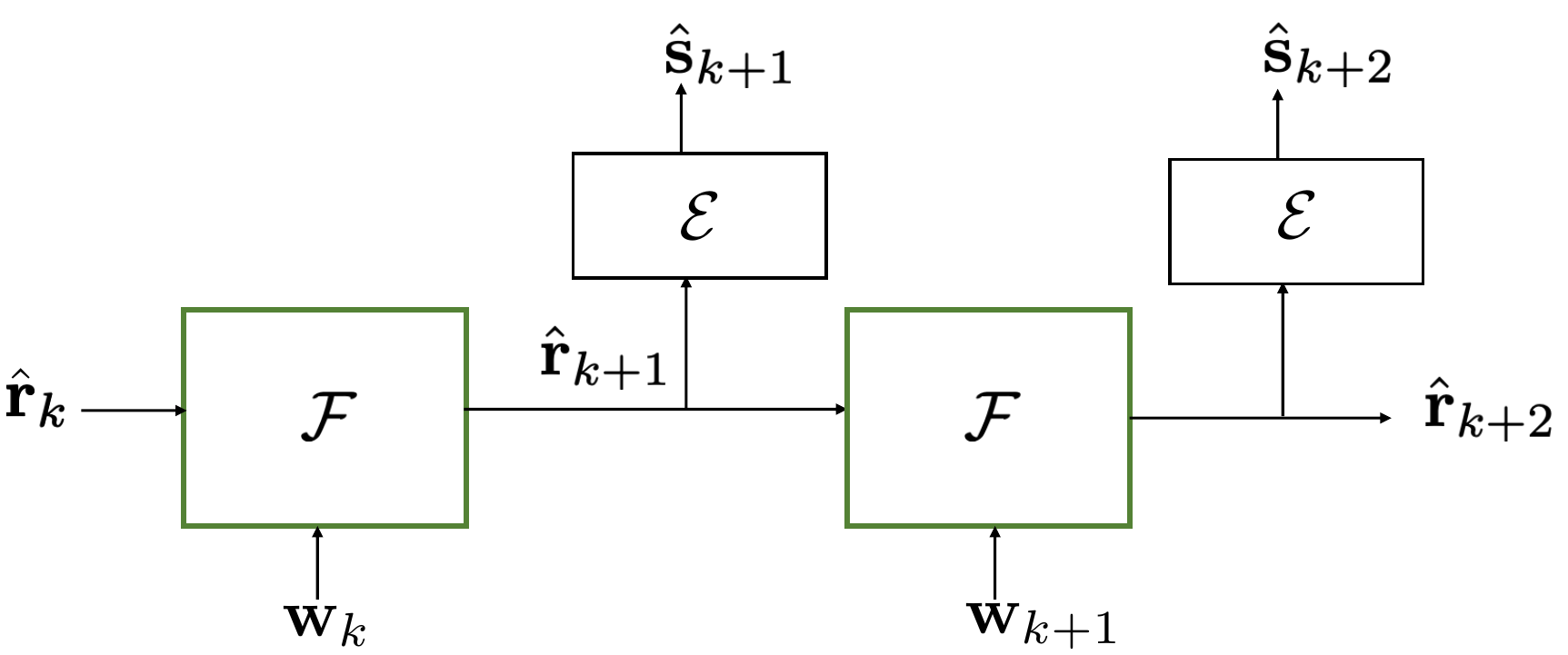}
    \caption{Illustration of repeated compositions of the time evolution operator to estimate the coordinates $\hat{\r}$ and $\hat{\s}$, for $SO_2$ and sulfate, respectively. The zonal wind coordinates $\w_k$ are input at each time step. The flow map $\mathcal F$  evolves the $SO_2$ and the operator $\mathcal E$ predicts the corresponding sulfate.}
  \label{fig:dnn_architecture}
\end{figure}

Our reduced model takes the initial $SO_2$ coordinate $\r_0$ and the time series of zonal wind coordinates $\{\w_k\}_{k=0}^{N_t}$ as input. The flow map~\eqref{eqn:forward_euler} is composed with itself $N_t$ times to produce a time series of approximate $SO_2$ coordinates $\{\hat{\r}_k\}_{k=1}^{N_t}$. The initial sulfate coordinates are determined by the initial $SO_2$ coordinates and $\mathcal E$ is applied at each time step $t_k$, $k\ge 1$, to compute sulfate coordinates $\{\hat{\s}_k\}_{k=1}^{N_t}$. The model architecture is illustrated in Figure~\ref{fig:dnn_architecture}.

The embedding of spatio-temporal structure and physical constraints in the model architecture is crucial to facilitate learning with limited data. However, the model architecture must be complemented with specialized a loss function to embed additional structure. We define the loss as the sum of misfits in both the $SO_2$ and sulfate predictions. Furthermore, to ensure stability in the time stepping, we consider a look ahead loss function which composes $\mathcal F$ with itself to predict over longer time horizons in the loss. Specifically, we define the loss function as
\begin{align}
\label{eqn:look_ahead_loss}
\sum\limits_{k=1}^{N_t} \sum\limits_{p=1}^{ \min{ \left( P,N_t-k \right) } } \vert \vert \r_{k+p} - \mathcal F^{[p]}(\r_k,\{\w_i\}_{i=k}^{k+p-1}) \vert \vert^2 + \vert \vert \s_{k+p} - \mathcal E(\mathcal F^{[p]}(\r_k,\{\w_i\}_{i=k}^{k+p-1})  ) \vert \vert^2
\end{align}
where $ \mathcal F^{[p]}(\r_k,\{\w_i\}_{i=k}^{k+p-1})$ denotes the composition of $\mathcal F$ to step from time $t_k$ to $t_{k+p}$, and $P$ is a tunable hyperparameter defining how many time steps we look ahead. A more detailed discussion of the forward Euler architecture~\eqref{eqn:forward_euler} and look ahead loss function~\eqref{eqn:look_ahead_loss} can be found in~\citep{Hart_2023}. We note the importance of the look ahead hyperparameter $P$ which encourages time stepping stability of the learned flow map. We also note that the loss function could be defined using only the $SO_2$ data. However, we observed that including the sulfate data added negligible computational cost and provided some benefit in generalization of the model.

\section{Inverse problem formulation}\label{sec:bayes_inv}

Given a trained flow map $\mathcal F$ and $SO_2$ to sulfate mapping $\mathcal E$, we define a mapping 
\begin{align*}
\mathcal G:\R^{3 N_{rbf}} \times \R^{N_w N_{rbf} N_t} \to \R^{3 N_{rbf} N_t}
\end{align*}
 from an initial volcanic sourced tagged $SO_2$ RBF coordinates $\r_0$ and time series of the zonal wind coordinates $\{\w_k\}_{k=0}^{N_t-1}$, to the time series of sulfate RBF coordinates, i.e. $\{ \hat{\s}_k \}_{k=1}^{N_t} = \mathcal G(\r_0,\{\w_k\}_{k=0}^{N_t-1})$.
  
We assume that only AOD observations are available and hence must model the mapping from source tagged sulfate to source tagged AOD. In earth system models, the AOD is computed point-wise in space and time using a model of scattering and absorption of light via atmospheric particulates. We can learn this mapping from the sourced tagged sulfate and AOD training data
 $$\{ \beta_v^{i,j},\rho_v^{i,j} \}_{i=1,j=1}^{N_e,N_s}.$$
  Specifically, we represent sulfate and AOD in the RBF basis and learn a mapping $\mathcal C: \R^{3 N_{rbf}} \to \R^{3 N_{rbf}}$ which inputs the source tagged sulfate RBF coordinates $\s_k$ and returns the source tagged AOD RBF coordinates $\q_k$.  
  
 The observable data (from satellite measurements) corresponds to point-wise evaluations of AOD at times $t_k$, $k=1,2,\dots,N_t$, and at $N_\text{obs}$ spatial locations in $\Omega_\text{lon}$. We define the operator $\mathcal B: \R^{3 N_{rbf}N_t} \to \R^{N_\text{obs} N_t}$ which, for each time step $t_k$, $k=1,2,\dots,N_t$, maps from AOD RBF coordinates to physical space via~\eqref{eqn:gaussian_basis_periodic} and evaluates the AOD at the $N_\text{obs}$ discrete observation points in space.
  
Composing all of these operators, we define
 \begin{align*}
 \mathcal A = \mathcal B \circ \mathcal C \circ \mathcal G: \R^{3 N_{rbf}} \times \R^{N_w N_{rbf} N_t} \to \R^{N_\text{obs} N_t}
\end{align*} 
which maps the initial $SO_2$ RBF coordinates and reduced zonal wind coordinates to the observable AOD at $N_\text{obs} N_t$ locations in space-time. We seek to compare predictions of $\mathcal A$ with observed AOD data $\d \in \R^{N_\text{obs} N_t}$ to estimate the initial $SO_2$. 

Uncertainty arises from a variety of processes: noise in AOD observations $\d$, background AOD not modeled in $\mathcal A$, and variability in zonal winds. We model noise in the AOD observations via a mean-zero Gaussian random vector $\mathbf{\epsilon} \in \R^{N_\text{obs} N_t}$ with covariance $\mathbf{\Sigma}_{\mathbf{\epsilon}}$. To incorporate the background AOD, we introduce a Gaussian random vector $\mathbf{\nu} \in \R^{N_\text{obs} N_t}$ whose mean $\mathbf{\mu}_{\mathbf{\nu}}$ and covariance $\mathbf{\Sigma}_{\mathbf{\nu}}$ are estimated from the AOD background data $\{ \rho_b^{i,j} \}_{i=1,j=1}^{N_e,N_s}$.

Then for initial $SO_2$ hyperparameters $\r_0$ and reduced zonal wind coordinates\\ $\w=(\w_0,\w_1,\dots,\w_{N_t-1})$, we  express the observed noisy AOD, $\d$, as
\begin{align*}
\d = \mathcal A(\r_0,\w) + \mathbf{\nu} + \epsilon .
\end{align*}
Assuming that the AOD background and observation noise are independent, if follows that $\d - \mathcal A(\r_0,\w)$ is normally distributed with mean $\mu_\mathbf{\nu}$ and covariance $\mathbf{\Sigma}_{\mathbf{\epsilon}} + \mathbf{\Sigma}_{\mathbf{\nu}}$.

Given a prior for $\r_0$ and $\w$, we apply Bayes rule to arrive at a joint posterior for $(\r_0,\w)$. The zonal wind $\w$ is uncertain due to the earth system's internal variability, but is not a quantity of interest to estimate. Rather we seek to account for zonal wind uncertainty in our estimate of $\r_0$. Accordingly, we take a Bayesian approximation error (BAE) approach~\citep{Kaipio_2013} by marginalizing over $\w$. Specifically, we consider
\begin{align*}
\d & = \mathcal A(\r_0,\w) + \mathbf{\nu} + \epsilon \\
& =  \mathbb{E}_\w \left[ \mathcal A(\r_0,\w) \right]  + \left( \mathcal A(\r_0,\w) - \mathbb{E}_\w \left[ \mathcal A(\r_0,\w) \right] \right)  + \mathbf{\nu} + \epsilon .
\end{align*}
We make the simplifying assumptions that $ \mathcal A(\r_0,\w) - \mathbb{E}_\w \left[ \mathcal A(\r_0,\w) \right] $ follows a Gaussian distribution, and that its covariance does not depend on $\r_0$. The latter assumption is reasonable as the variability due to atmospheric winds is known to be chaotic and hence will have a weak dependence on the aerosol mass. The Gaussian assumption does not have theoretical justification. However, this assumption is common in the BAE literature and has been observed to be reasonable in many applications. It results in a convenient expression for the modified likelihood which is both computable and interpretable.

We compute samples of $\mathcal A(\r_0,\w) - \mathbb{E}_\w \left[ \mathcal A(\r_0,\w) \right]$ by drawing prior samples from $(\r_0,\w)$ and propagating them through $\mathcal A$. Ignoring its dependence on $\r_0$, we fit a Gaussian distribution to these samples. This yields a mean $\mathbf{\mu}_\text{BAE}$ and covariance $\mathbf{\Sigma}_\text{BAE}$ which models variability due to the zonal winds. Then we have that
\begin{align*}
\mathbf{\eta} = \left( \mathcal A(\r_0,\w) - \mathbb{E}_\w \left[ \mathcal A(\r_0,\w) \right] \right)  + \mathbf{\nu} + \epsilon 
\end{align*}
is Gaussian with mean $\mathbf{\mu}=\mathbf{\mu}_{\mathbf{\nu}} + \mathbf{\mu}_\text{BAE}$ and covariance $\mathbf{\Sigma} = \mathbf{\Sigma}_{\mathbf{\epsilon}} + \mathbf{\Sigma}_{\mathbf{\nu}} + \mathbf{\Sigma}_\text{BAE}$. Hence, 
\begin{align*}
\d & = \mathbb{E}_\w \left[ \mathcal A(\r_0,\w) \right]+ \mathbf{\eta}
\end{align*}
and we express the posterior PDF for $\r_0$ as
\begin{align}
\label{eqn:bae_post}
\pi_\text{post}(\r_0 \vert \d) \propto \pi_\text{like}(\d \vert \r_0) \pi_\text{prior}(\r_0)
\end{align}
where
\begin{align}
\label{eqn:likelihood}
\log \left( \pi_\text{like}(\d \vert \r_0) \right) = -\frac{1}{2} \left( \mathbb{E}_\w \left[ \mathcal A(\r_0,\w) \right] + \mathbf{\mu} -  \d \right)^T \mathbf{\Sigma}^{-1} \left( \mathbb{E}_\w \left[ \mathcal A(\r_0,\w) \right] + \mathbf{\mu} -  \d \right) 
\end{align}
and $\pi_\text{prior}$ is the prior PDF for $\r_0$. 

We emphasize the following points regarding~\eqref{eqn:likelihood}. First, the background AOD is accounted for in $\mathbf{\mu}$ and $\mathbf{\Sigma}$ due to the source tag separating the volcanic and background aerosols. This allows us to avoid modeling dynamics of the small scale processes such as industrial emissions or dust storms while still accounting for them in the inverse problem. Second, internal climate variability that manifests itself in wind uncertainty is accommodated in~\eqref{eqn:likelihood} through: (i) computing the average observable AOD by taking an expectation over the zonal winds $\w$, and (ii) weighting the data misfit with a covariance $\mathbf{\Sigma}$ whose magnitude depends on the magnitude of AOD variability due to wind uncertainty (measured in $\mathbf{\Sigma}_\text{BAE}$). Our assumptions about the statistics of $\mathcal A(\r_0,\w) - \mathbb{E}_\w \left[ \mathcal A(\r_0,\w) \right]$ enabled this convenient expression for the likelihood in terms of a bias correction $\mathbf{\mu}=\mathbf{\mu}_{\mathbf{\nu}} + \mathbf{\mu}_\text{BAE}$ and uncertainty weighting $\mathbf{\Sigma} = \mathbf{\Sigma}_{\mathbf{\epsilon}} + \mathbf{\Sigma}_{\mathbf{\nu}} + \mathbf{\Sigma}_\text{BAE}$.

\begin{figure}[h]
\centering
\includegraphics[width=0.99\textwidth]{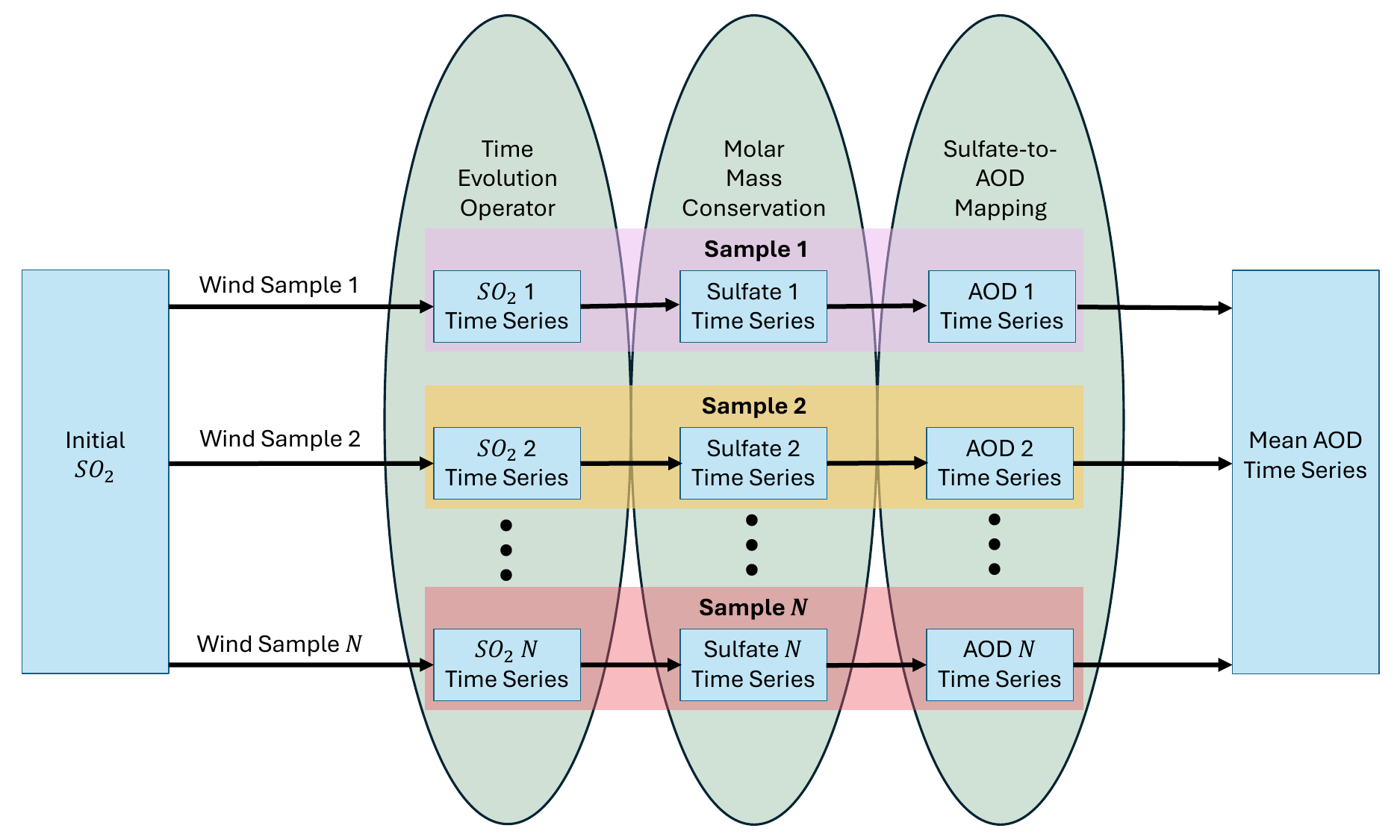}
    \caption{Depiction of the forward model used in our inversion. The leftmost box corresponds to $\r_0$ and the rightmost box is a sample estimator of $\mathbb{E}_\w \left[ \mathcal A(\r_0,\w) \right]$. The inverse problem seeks to determine $\r_0$ such that $\d = \mathbb{E}_\w \left[ \mathcal A(\r_0,\w) \right]+ \mathbf{\eta}$, where $\d$ is the observed AOD data and $\eta$ is random noise which incorporates an estimate of background AOD and internal variability from the climate system. The $N$ samples correspond to taking one $\r_0$ and sampling $N$ $\w$'s to approximate  $\mathbb{E}_\w \left[ \mathcal A(\r_0,\w) \right]$ via $N$ samples.}
  \label{fig:inversion_diagram}
\end{figure}

Figure~\ref{fig:inversion_diagram} summarizes how the models are combined in our inversion framework. The leftmost box in Figure~\ref{fig:inversion_diagram} corresponds to the initial $SO_2$ RBF coordinates which are being estimated. For a given initial $SO_2$ plume, the figure shows how the models are composed to propagate $N_w$ reduced wind samples to produce $N_w$ time series of AOD predictions, which are averaged in the rightmost box of Figure~\ref{fig:inversion_diagram}. The inverse problem seeks an initial $SO_2$ (leftmost in the figure) such that the average AOD prediction (rightmost in the figure) matches the observed AOD data, with a correction for background AOD and weighting to account for noise and variability. 

\section{Numerical results} \label{sec:numerical_results}

We demonstrate our proposed framework using the following datasets. The training data consists of $N_e=7$ ensemble members. For each ensemble member, $N_s=5$ simulations were generated which correspond to  $SO_2$ injection source magnitudes of $3,5,7,13,$ and $15$ teragrams (Tg). This gives a total of $N_eN_s=35$ simulations used in the training set, where each simulation includes the variables in (\ref{eqn:raw_data}). Five simulations from an eighth ensemble member with source magnitudes of $3,5,7,13,$ and $15$ Tg are held out as a validation set. The test set consists of two simulations from a ninth and tenth ensemble member with source magnitude $10$ Tg, similar to the Mount Pinatubo eruption magnitude. This setup ensures that the numerical results are derived from ensembles and source magnitudes that were not included in the training set, as  summarized in Table~\ref{tab:dataset_overview}. Our results consider daily data (a time resolution of $24$ hours) and a time horizon of 10 days, i.e. $N_t=9$.  

\begin{table}[!ht]
\centering
\includegraphics[width=0.99\textwidth]{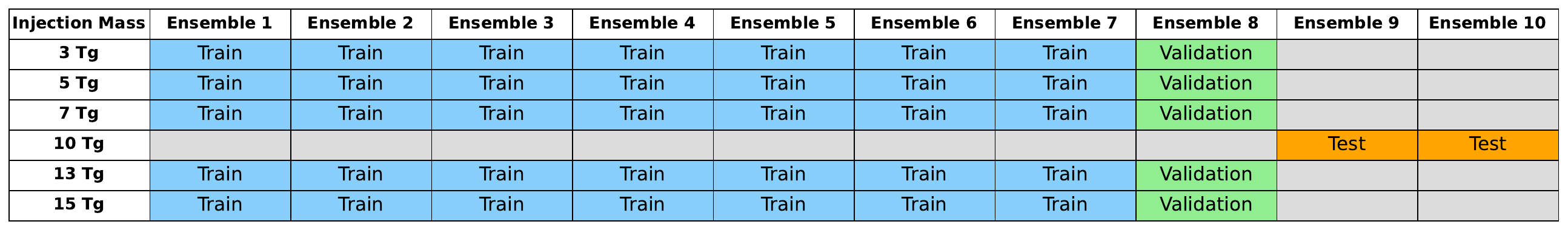}
\captionsetup{justification=centering}
\caption{Overview of the simulations included in the training (blue), validation (green), and test (orange) sets. }
\label{tab:dataset_overview}
\end{table}

\begin{figure}[h]
\centering
\includegraphics[width=0.99\textwidth]{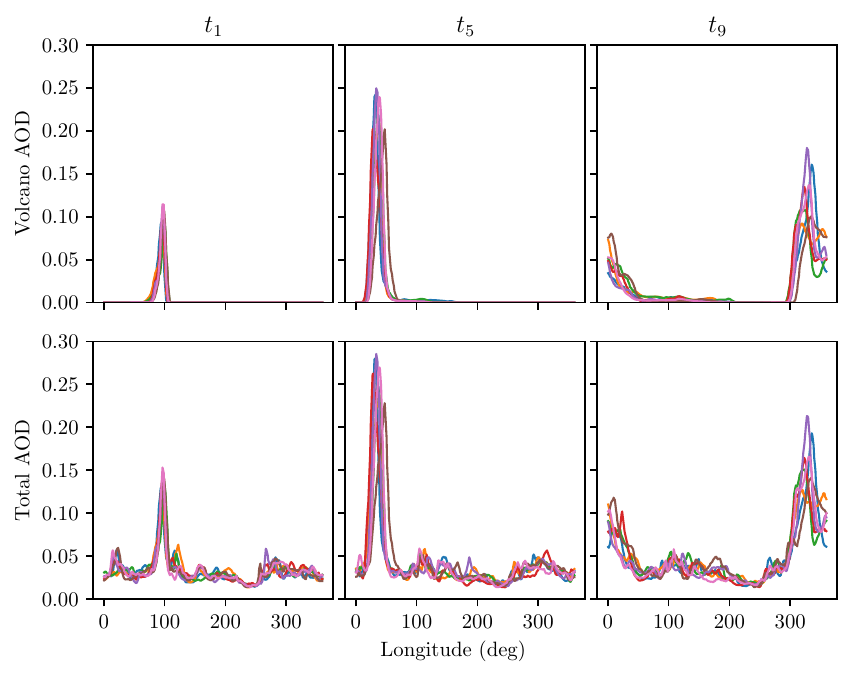}
    \caption{Top: source tagged volcanic AOD; bottom: volcanic plus background AOD. From left to right: days $1,5,$ and $9$. Each panel displays $7$ ensemble members with a fixed injection magnitude.}
  \label{fig:simulation_data}
\end{figure}

The observed data informing the inverse problem is the sum of volcanic and background AOD.  Figure~\ref{fig:simulation_data} highlights the benefit of using source tagged data for training the time evolution operator by comparing the volcanic source tagged AOD with the AOD due to background aerosols. Each panel in the figure displays 7 curves corresponding to the $N_e=7$ training ensembles with a fixed source magnitude\footnote{The source magnitude in Figure~\ref{fig:simulation_data} is $7$ Tg.}. The top row displays the source tagged AOD corresponding to the volcanic aerosols, $\rho_v$, and the bottom row displays the total AOD (volcanic and background), i.e. $\rho_v + \rho_b$. Distinguishing the volcanic AOD from the background noise becomes harder over time due to both the variability across ensembles as demonstrated by comparing days $t_0$ and $t_9$ in the top row, and  the dispersion of the plume over time which reduces the signal to noise ratio, as demonstrated by comparing days  $t_5$ and $t_9$ in the bottom row.

In the subsections below, we present numerical results mirroring the progression of the approach in Sections~\ref{sec:spatial_reduction},~\ref{sec:time_evol}, and~\ref{sec:bayes_inv}.

\subsection{Spatial dimension reduction}
\subsubsection{Aerosol dimension reduction}
With time horizon of $N_t=9$ days, it is sufficient to use a single radial basis function, i.e. $N_{rbf}=1$, to represent the plume (see Figure~\ref{fig:simulation_data}). By day $t_9$, the plume is asymmetric and heavy tailed, so there is noticeable error in using a single Gaussian basis function. However, we justify our choice of not including additional RBFs by noting the error is on a comparable order of magnitude as the background aerosol and hence is negligible in the inverse problem. By using this low-dimensional embedding of the spatial field into a $3$ dimensional space, we ensure that the time evolution operator can be trained effectively even with a small dataset of only $35$ time series.

\begin{figure}[h]
\centering
  \includegraphics[width=0.49\textwidth]{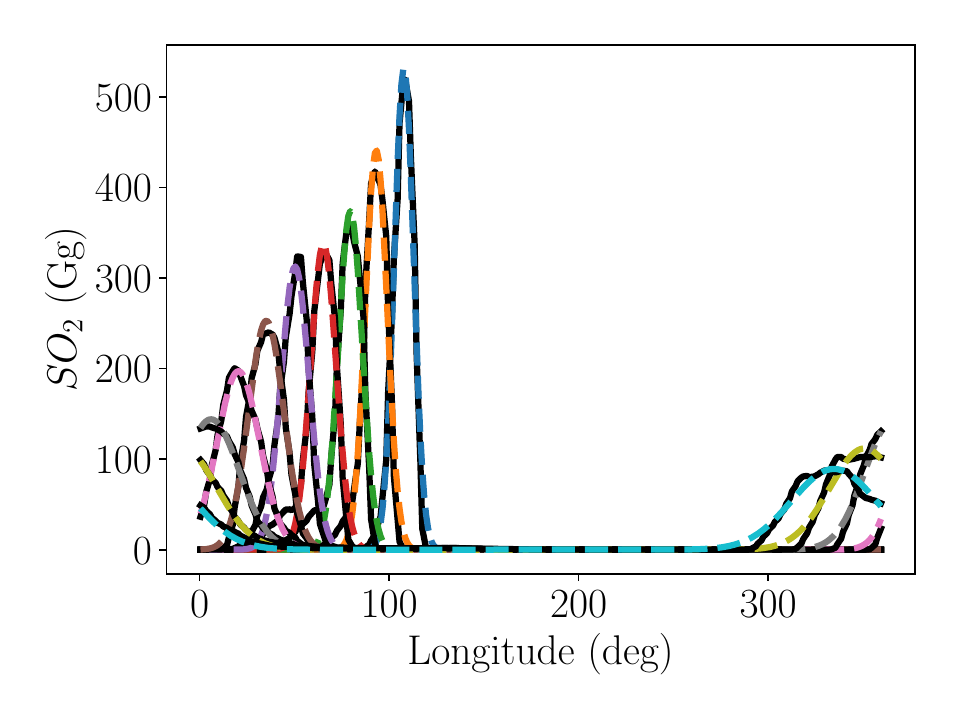}
    \includegraphics[width=0.49\textwidth]{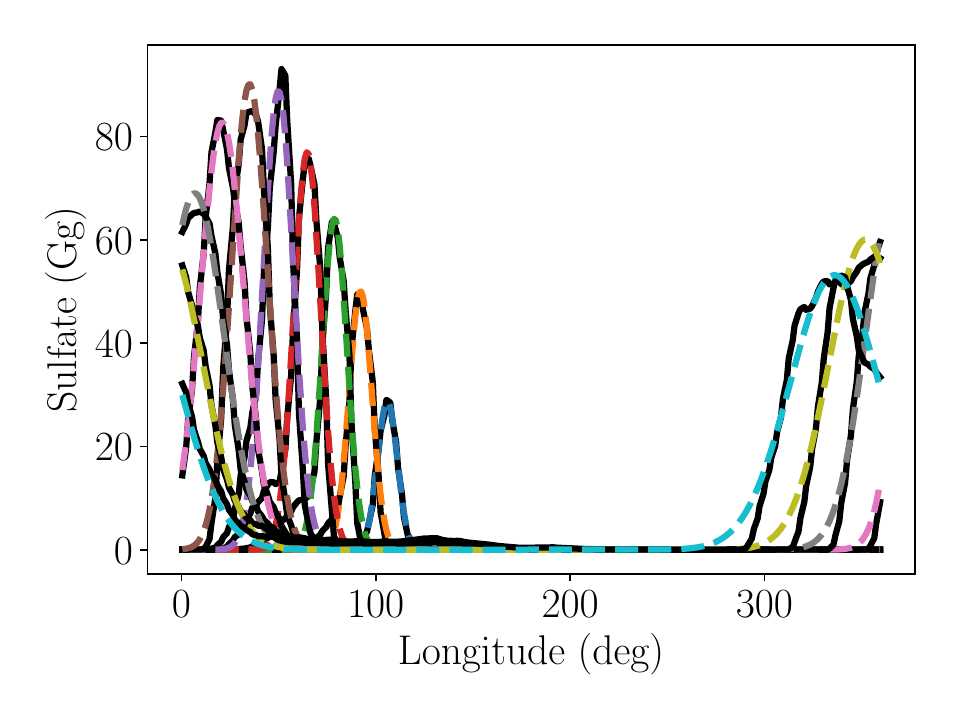}
    \caption{RBF fit for volcanic $SO_2$ (left) and sulfate (right) in gigagrams (Gg). The solid black lines show the raw data and the colored broken lines show the RBF fit, with a different color for each time step . Note that $t_0$ corresponds to the blue line, and the plume advects westward.}
  \label{fig:rbf_fit}
\end{figure}

We fit the RBF using a block-coordinate-descent optimization algorithm which alternates between a gradient descent step to update the scale and shape hyperparameters and a linear least squares solve to update the coefficient hyperparameter.  Figure~\ref{fig:rbf_fit} shows the RBF fit for both the volcanic $SO_2$ (left) and sulfate (right) variables. We observe the trend that the $SO_2$ mass decreases as the sulfate mass increases (as functions of time). The maximum sulfate value initially increases and subsequently decreases as the plume diffuses in space. The $SO_2$ and sulfate RBFs have nearly identical center and shape hyperparameters. The center is monotonically decreasing (moving right to left and then through the boundary at 0$^\circ$ longitude, the Greenwich prime meridian) and the shape hyperparameter is monotonically decreasing as the plume diffuses in space. The near identical shape and center hyperparameters motivate our flow map model which evolves the $SO_2$ and estimates the sulfate using the same center and shape hyperparameters as the $SO_2$ RBF representation.

\begin{figure}[h]
\centering
  \includegraphics[width=0.49\textwidth]{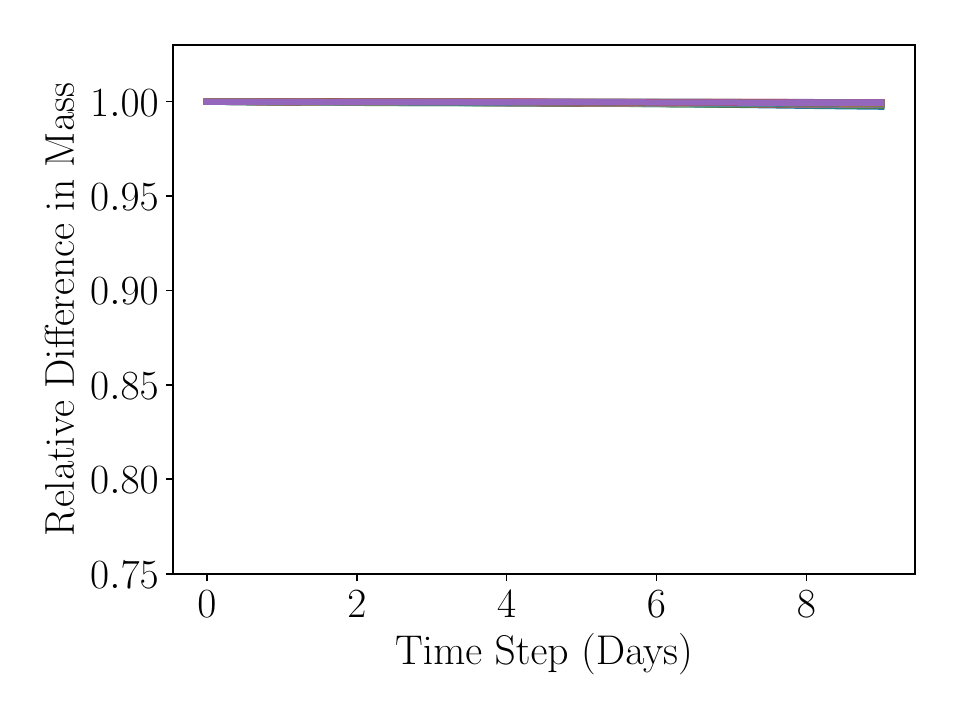}
    \includegraphics[width=0.49\textwidth]{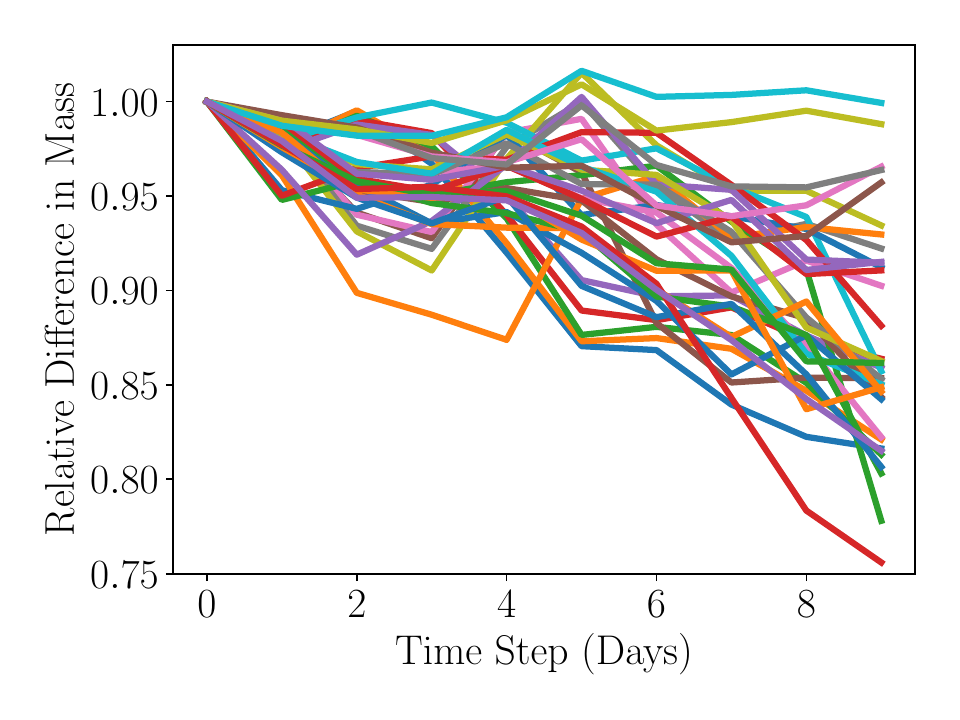}
    \caption{Relative loss in total mass of sulfur as a function of time in the raw data (left panel) and the RBF approximation (right panel). Each panel has $35$ curves corresponding to the training set simulations.}
  \label{fig:rbf_mass_lost}
\end{figure}

Due to the inability of the Gaussian RBF basis function to capture heavy tails in the plume, the conservation of mass equation~\eqref{eqn:conservation} is not satisfied by the RBF approximation. Figure~\ref{fig:rbf_mass_lost} shows the relative loss in total source tagged $SO_2$ mass as a function of time in the raw data (left panel) and the RBF approximation (right panel). Specifically, letting $S:[0,T] \to \R$ denote the total mass of source tagged volcanic $SO_2$ in the atmosphere as a function of time, Figure~\ref{fig:rbf_mass_lost} displays $S(t)/S(0)$ for all simulations in the training set.  The simulation curves in the left panel correspond to the constant function $S(t)/S(0)=1$ (with some small numerical noise) since mass is conserved. In contrast, the curves in the right panel disperse since the RBF approximation does not preserve mass. We observe a loss in mass between $5\%$ and $15\%$ for the majority of the time series. This indicates that additional RBF basis functions may be required to improve the accuracy of the reduced order model we seek to learn. However, in practice we use a single RBF basis function as the subsequent results are sufficient in spite of this error in mass conservation. We posit that this lack of mass conservation is acceptable for two reasons. First, the network architecture enforces conservation of mass and its loss function incorporates both $SO_2$ and sulfate data, so the network cannot overfit to the loss of RBF mass in the training data. Second, the inverse problem is most informed by the earlier time steps in which the loss of mass is smaller than the background aerosol noise.

\subsubsection{Wind dimension reduction}
\begin{figure}[h]
\centering
    \includegraphics[width=0.99\textwidth]{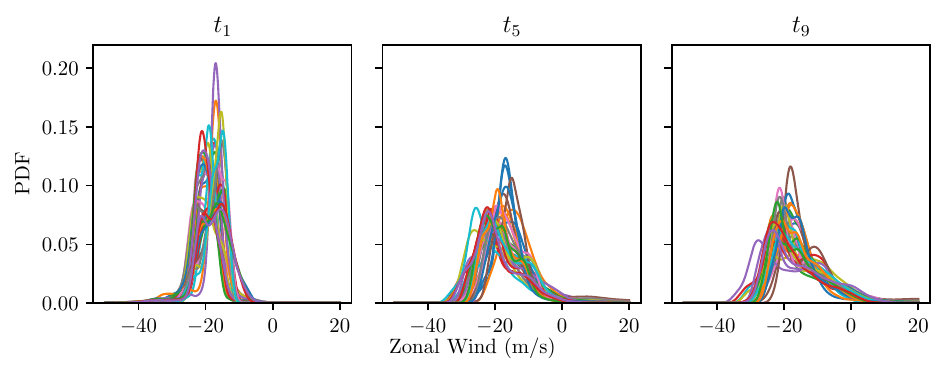}
    \caption{Time snapshots of zonal wind PDFs for days $1$ (left), $5$ (center), and $9$ (right). Each panel has $35$ PDFs corresponding to the training set simulations. This 1D representation of the zonal wind is computed using samples of the 3D wind field weighted by the magnitude of $SO_2$ present in the atmosphere.}
  \label{fig:zonal_wind_pdfs}
\end{figure}

To embed the zonal wind data, we select threshold $\tau_{SO_2}=100$ grams to truncate the wind data to the region of the atmosphere where the mass of $SO_2$ is greater than $\tau_{SO_2}$ as defined in \eqref{eqn:wind_tau}. Weighting by the RBF basis function at each time step \eqref{eqn:wind_rbf}, we generate probability density functions (PDFs) corresponding to the distribution of the zonal winds localized about the plume at each time step. This gives a total of $N_e N_s N_t=(7)(5)(10)=350$ PDFs corresponding to the zonal winds at each time step in each training set simulation. Snapshots of these zonal wind PDFs are shown in Figure~\ref{fig:zonal_wind_pdfs} with the left, center, and right panels corresponding to days $1,5,$ and $9$. Each panel has $35$ PDFs corresponding to the $N_eN_s=35$ training set simulations. We observe that the width of the PDFs, i.e. the level of uncertainty in the zonal wind, is increasing over time. This is a result of the ensembles drifting further apart as time evolves.

\begin{figure}[h]
\centering
    \includegraphics[width=0.49\textwidth]{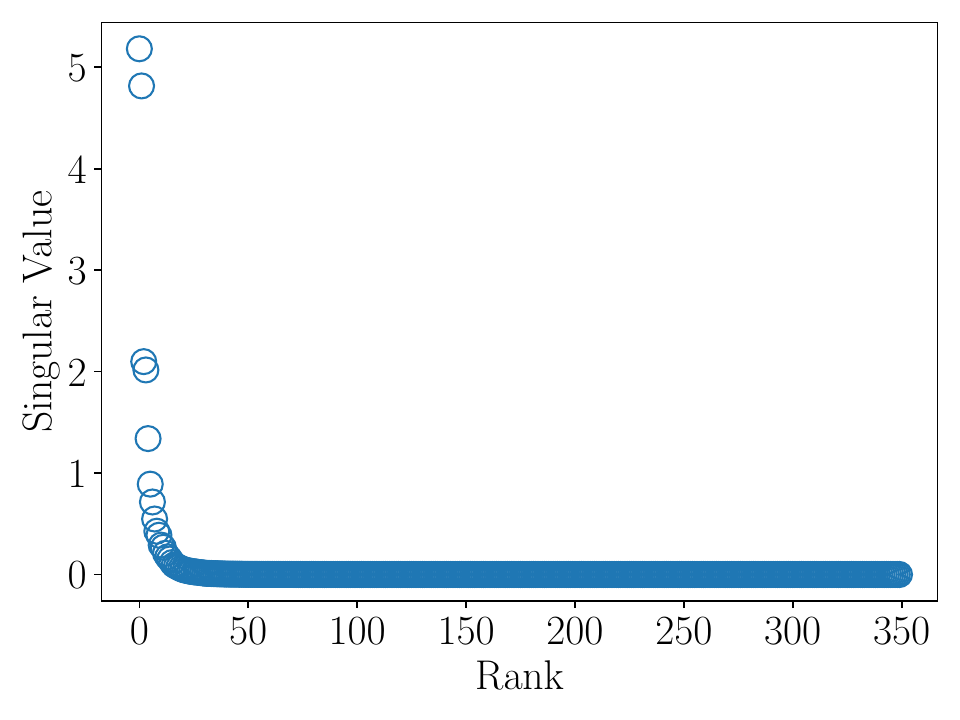}
  \includegraphics[width=0.49\textwidth]{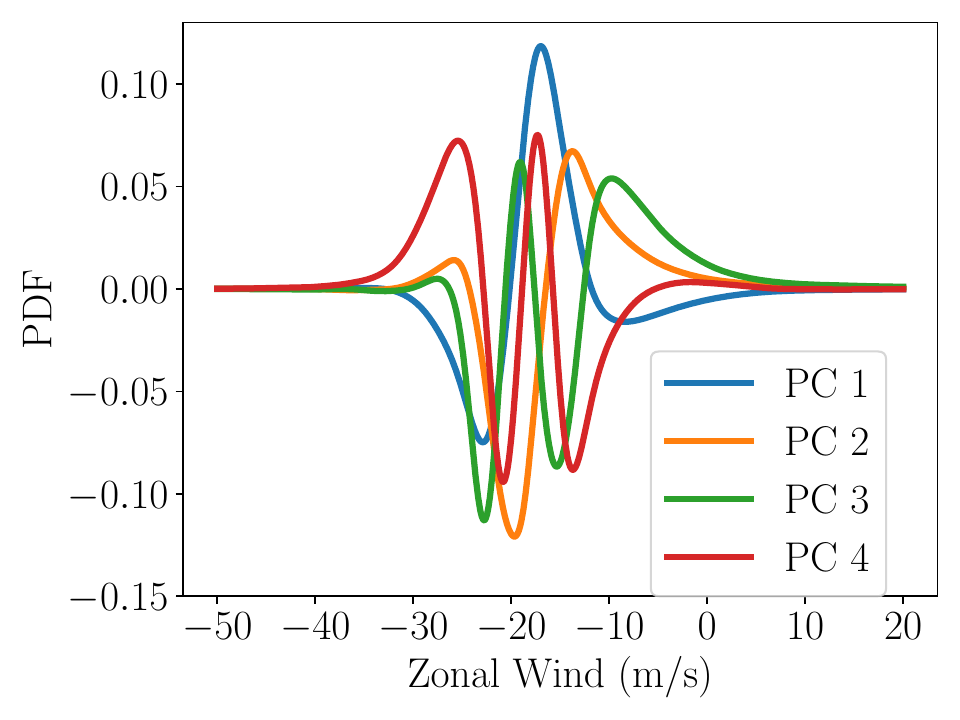}
    \caption{Left: singular values of the zonal wind PDF data matrix. Right: leading principle components (PCs) of the zonal wind PDF data matrix.}
  \label{fig:zonal_wind_pca}
\end{figure}

We evaluate these $350$ PDFs on a grid of $1000$ zonal wind points to form a $1000 \times 350$ matrix. Principle component analysis is applied to this matrix after centering by subtracting the mean from each column. The left panel of Figure~\ref{fig:zonal_wind_pca} displays the singular values of the centered data matrix. The right panel of the figure shows the four leading singular vectors, known as the principle components (PCs). We choose a truncation rank of $N_w=4$ based upon the spectral characteristics and properties of the principle components displayed in Figure~\ref{fig:zonal_wind_pca}.  Note that there is a nontrivial truncation as the $5^{th}$ singular value is roughly $1/4$ of the $1^{st}$ singular value's magnitude. However, in our context, this loss of information is helpful as it corresponds to removing higher frequency wind variations so that the reduced wind coordinates capture the larger scale variations. In~\ref{appendix:wind_modes}, we further justify this choice by analyzing the validation set prediction error for models with various truncation ranks $N_w$.

\subsection{Time evolution operator}

After dimension reduction, we have $35$ time series of $SO_2$ RBF coordinates in $\R^3$ and zonal wind principal component coordinates in $\R^4$. We train a neural network time evolution operator $\mathcal N:\R^3 \times \R^4 \to \R^3$ with the architecture 
\begin{align}
\label{eqn:NN_arch}
\mathcal N(x,a,c,\w) = T^{-1} \left( \sigma \left(\mathbf{L} \left( \begin{bmatrix} T(x,a,c) \\ \w_k   \end{bmatrix} \right) \right) \right)
\end{align}
where $(x,a,c) \in \R^3$ are the center, scale, and coefficient for the $SO_2$ RBF approximation, $\w_k \in \R^4$ are the zonal wind principal component coordinates, $\mathbf{L}$ represents a single fully connected linear layer,  $\sigma:\R^3 \to \R^3$ is the element-wise activation function $\sigma_i(y) = \min\{0,y\}$, $i=1,2,3$, and $T:\R^3 \to \R^3$ is the coordinate transformation
\begin{align*}
T(x,a,c) = \left(x,a,\frac{c}{a} \right).
\end{align*}

The architecture defined in \eqref{eqn:NN_arch} ensures that the flow map predictions for unseen data preserves the known physical properties of the aerosol transport and chemistry. We accomplish this by imposing that the total mass of $SO_2$ and the RBF center and scale hyperparameters are monotonically decreasing over time as this corresponds to the physical processes of advection and diffusion. We impose monotonicity in the network architecture using the activation function $\sigma$. The coordinate transformation $T$ maps the RBF coefficient $c$ to the quotient $\frac{c}{a}$, a scalar multiple of the total mass of $SO_2$ that arises from integration of the radial basis function. The inverse coordinate transformation, $T^{-1}$, is applied to the network output. 

The choice of a single linear layer $\mathbf{L}$ was made by an analysis wherein we consider architectures of the form~\eqref{eqn:NN_arch}, as well as deeper architectures with additional hidden layers. Instead of determining the optimal architecture from a validation set prediction error, we considered the prediction of the network on a larger set of inputs and measured how well it capture the $SO_2$ to sulfate reaction rate. Rather than being confined to a small validation set, this metric explores the model predictions over the full range of inputs that may be seen in the inverse problem. This analysis, which lead to our choice of a single linear layer, is detailed in~\ref{appendix:depth_analysis}. 

\begin{figure}[h]
\centering
  \includegraphics[width=0.49\textwidth]{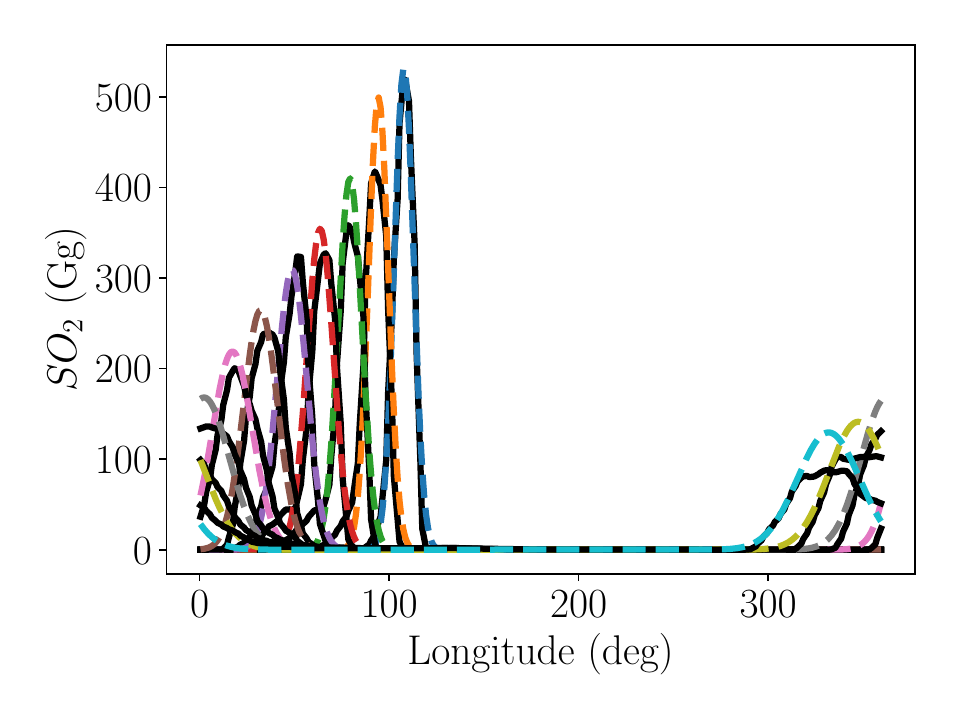}
    \includegraphics[width=0.49\textwidth]{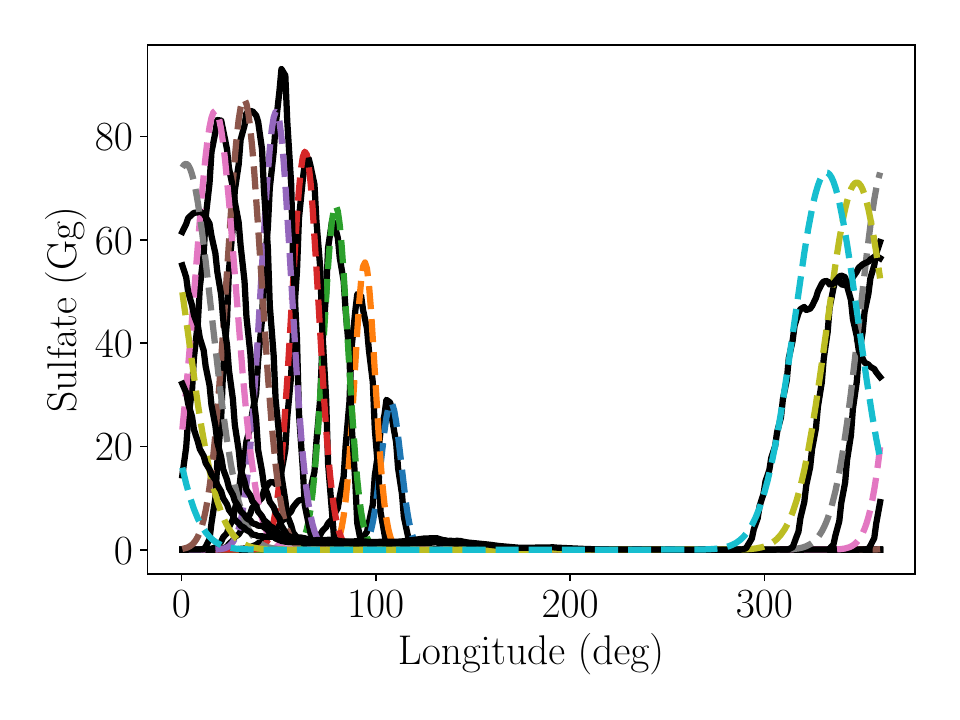}
    \caption{Validation set prediction for volcanic $SO_2$ (left) and sulfate (right) in Gigagrams (Gg). The solid black lines show the raw data and the colored broken lines show the prediction, with a different color for each time step. Note that $t_0$ corresponds to the blue line, and the plume advects westward.}
  \label{fig:network_val_pred}
\end{figure}

The dense feed forward neural network is trained using batch gradient descent, where each batch includes the training simulations from 2-3 ensembles over all source magnitudes. Figure~\ref{fig:network_val_pred} displays the validation set prediction of the volcanic $SO_2$ (left) and sulfate (right) variables. The raw data is given by the solid black lines and the prediction is given by the colored broken lines, with the colors distinguishing the time steps.

Given the time evolution operator~\eqref{eqn:NN_arch} to predict the $SO_2$ trajectory, we compute the sulfate plume as each time step by setting the sulfate RBF center and shape hyperparameter equal to that of the $SO_2$, and the coefficient is computed via the conservation of mass equation~\eqref{eqn:conservation}.

To map the sulfate RBF coordinates  to AOD RBF coordinates, a separate linear model  $\mathbf{L}_\text{AOD} \in \R^{3 \times 3}$ is fit  to the training data using ordinary least squares.  We were able to deploy a simple model for this step as the complexity of the problem was significantly reduced thanks to source tagging and the use of RBF hyperparameters rather than the full spatial dimensions.

\subsection{Inverse problem formulation}

\begin{figure}[h]
\begin{subfigure}
\centering
  \includegraphics[width=0.32\textwidth]{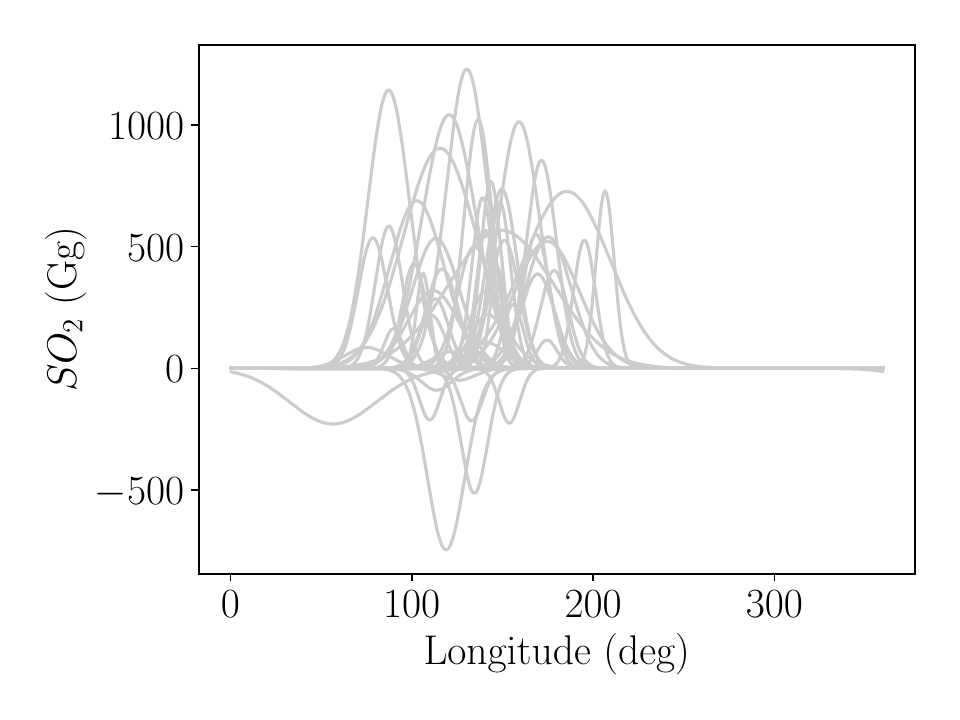}
    \includegraphics[width=0.32\textwidth]{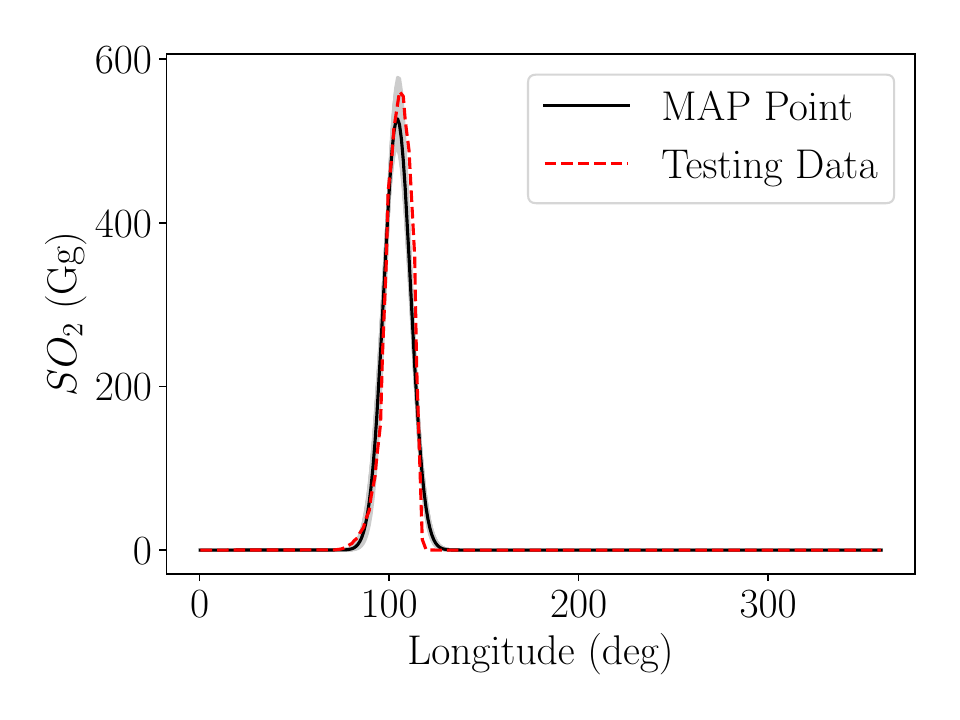}
        \includegraphics[width=0.32\textwidth]{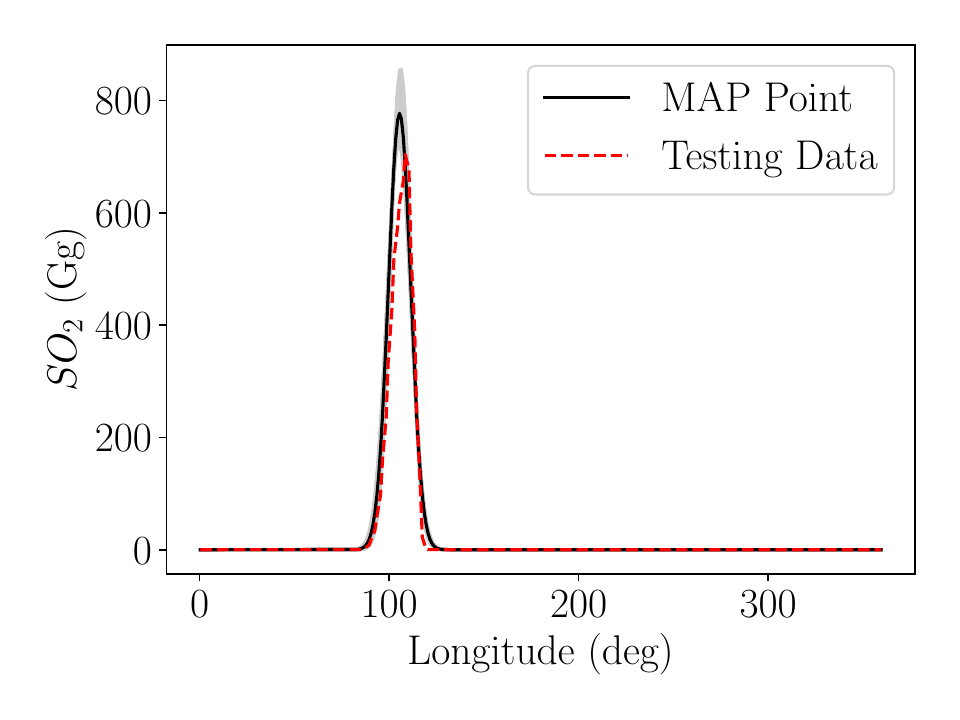}
  \end{subfigure}
  \caption{
  Inverse optimization results on two test ensemble simulations.  Left: prior samples of the initial $SO_2$ plume. Center (ensemble 9) and Right (ensemble 10): the MAP point and approximate posterior samples of the initial $SO_2$ plume (in grey), with the test data overlaid to demonstrate accuracy.}
  \label{fig:prior_and_post_samples}
\end{figure}

We specify a Gaussian prior, $\pi_\text{prior}(\r_0)$, on the initial $SO_2$ RBF coordinates. The left panel of Figure~\ref{fig:prior_and_post_samples} displays samples from the prior mapped from the RBF coordinate space back onto the physical space. We intentionally chose a prior with large variance to evaluate how much is learned from the data. The negative values are a result of the Gaussian assumption and large variance which results in some prior samples being physically infeasible. 

We assume a mean zero Gaussian noise model whose covariance is a multiple of the identity matrix, $\sigma_\text{noise}^2 \mathbf{I}$, where $\sigma_\text{noise}=0.01$ is approximately the magnitude of AOD measurement noise. The background AOD statistics $\mathbf{\mu}_{\mathbf{\nu}}$ and $\mathbf{\Sigma}_{\mathbf{\nu}}$ are computed from the $35$ training set time series for the background AOD variable. The BAE correction mean and covariance $\mathbf{\mu}_\text{BAE}$ and $\mathbf{\Sigma}_\text{BAE}$ are also determined using the $35$ simulations in the training set. Specifically, using our learned time evolution operator, we forward propagate the $SO_2$ prior mean using the $35$ wind field samples and compute the empirical mean and covariance of the samples.

We construct observational data by extracting the AOD (volcanic and background) from the test set and contaminating the data with mean zero additive Gaussian noise whose standard derivation is $0.012$ (chosen to be comparable but not equal to the noise model in the likelihood function). Numerical optimization is used to compute the maximum a posteriori probability (MAP) point of the posterior~\eqref{eqn:bae_post}. To perform this optimization, we use efficient derivative-based optimization algorithms in the Rapid Optimization Library~\cite{ROL-website}. Derivatives are computed by implementing the time evolution operator~\eqref{eqn:forward_euler} in TensorFlow and leveraging its algorithmic differentiation capability. The details of this approach are similar to those described in~\cite{Hart_2023}.

 Approximate posterior samples are computed using a Laplace approximation of the posterior. That is, we take a Gaussian approximation of the posterior with mean given by the MAP point and covariance given by the inverse Hessian of the negative log likelihood. The center and right panels of Figure~\ref{fig:prior_and_post_samples} displays the posterior MAP point for the two test simulations, approximate samples (given by the grey shading), and the ground true sources from the test set. We note that the optimization problem to determine the MAP point is non-convex and hence possess multiple local minima. We ran the optimization multiple times from different random initializations and chose the solution that attained the greatest likelihood value.

Comparing the two test sets, we observe greater accuracy estimating the $SO_2$ source for ensemble 9 (center panel of Figure~\ref{fig:prior_and_post_samples}) in comparison to ensemble 10 (right panel of Figure~\ref{fig:prior_and_post_samples}). The $\ell_2$ relative error between the MAP point and test data is $18 \%$ for ensemble 9 and $32 \%$ for ensemble 10. Both test sets used the same $SO_2$ injection but differ in their wind fields as a result of ensemble variability. To understand the wind variability, we compare the reduced wind coordinates $\mathbf{w} \in \R^{40}$ from ensembles 9 and 10 with the reduced wind coordinates from the training set. Note that the test set wind data is never used in our analysis as we do not assume precise knowledge of the stratospheric winds in observations, but it is available for this error analysis since the observational data was synthesized from simulations. To measure the distance between the training and test set reduced winds, we compute the Mahalanobis distance~\cite{maupin2018validation} between the training set reduced wind coordinates $\{ \mathbf{w}^i \}_{i=1}^{35} \subset \R^{40}$ and each test set reduced wind coordinate, $\mathbf{w}^{ens09}, \mathbf{w}^{ens10} \subset \R^{40}$. Specifically, we compute the empirical mean $\mathbf{\mu}_{\mathbf{w}}$ and covariance $\mathbf{\Sigma}_{\mathbf{w}}$ from $\{ \mathbf{w}^i \}_{i=1}^{35} \subset \R^{40}$ and the Mahalanobis distance
\begin{align*}
d^{ensX} = \sqrt{ \left( \mathbf{w}^{ensX} - \mathbf{\mu}_{\mathbf{w}} \right) \mathbf{\Sigma}_{\mathbf{w}}^{\dagger} \left( \mathbf{w}^{ensX} - \mathbf{\mu}_{\mathbf{w}} \right)  },
\end{align*}
for $ens09$ and $ens10$, where the distance from $\mathbf{\mu}_{\mathbf{w}}$ is weighted by the pseudo-inverse of the empirical covariance since $\mathbf{\Sigma}_{\mathbf{w}}$ has rank $34$ (one less than the number of training set simulations). We have $d^{ens09}=1530$ and $d^{ens10}=6372$. In relative terms, ensemble 10 is approximate 4 times further away from the training data then ensemble 9. This is consistent with the MAP point errors where our $SO_2$ estimation error for ensemble 10 is nearly double the error for ensemble 9. This confirms the unsurprising fact that having a training set close to the actual stratospheric winds from the observational period is crucial to achieve accurate aerosol estimates. 

\begin{figure}[h]
\centering
  \includegraphics[width=0.99\textwidth]{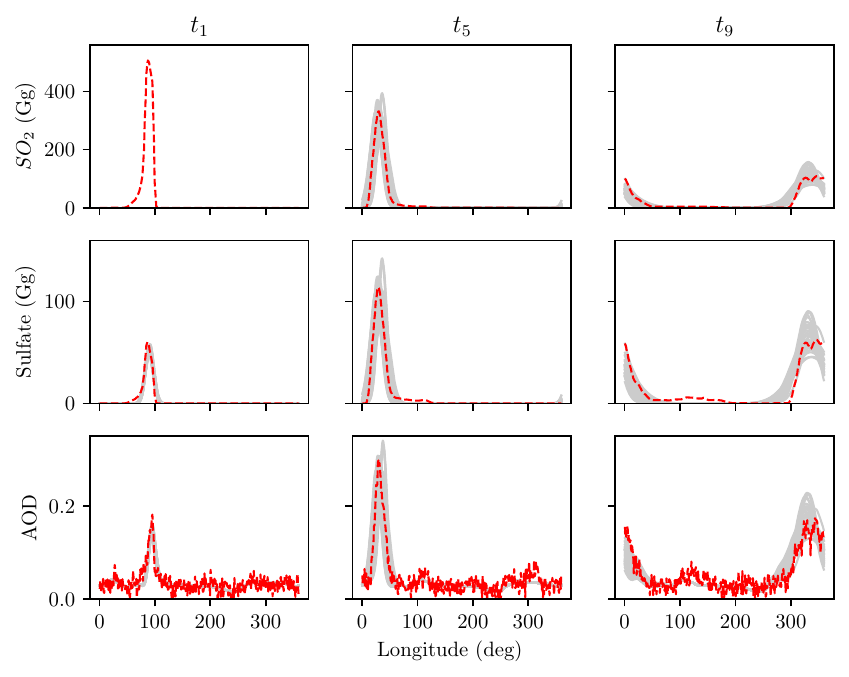}
    \caption{Predictions of $SO_2$ (top row), sulfate (middle row), and AOD (bottom row), alongside the Ensemble 9 test data (given by red broken curves), at day 1 (left), day 5 (middle), and day 9 (right). There are $35$ grey curves in each panel corresponding to the model prediction from the $35$ training ensemble zonal winds.}
  \label{fig:state_pred_at_map}
\end{figure}

 Figure~\ref{fig:state_pred_at_map} displays the $SO_2$ (top row), sulfate (middle row), and AOD (bottom row) predictions of the reduced model for Ensemble 9, alongside the test data which it is seeking to match. Three time snapshots, day 1 (left), day 5 (middle), and day 9 (right) are shown to illustrate the predictions time dependence. In each plot, there are $35$ grey curves corresponding to the model prediction using the wind fields from the $35$ training set simulations. By maximizing the likelihood \eqref{eqn:likelihood}, the average of the AOD predictions closely matches the observed data. We observe how uncertainty increases as a function of time. This is a result of the zonal wind variability increasing over time. The $SO_2$ and sulfate fields are less noisy compared to the AOD since they correspond to volcanic source tagged data, while the AOD fields include the background aerosol and measurement noise. 

\section{Conclusion} \label{sec:conclusion}

Estimating stratospheric aerosol sources is challenging. Internal variability in the climate system makes it difficult to learn a model for plume evolution since the operator must account for both variations in the source injection and the atmospheric winds. This challenge is compounded by the presence of background aerosols which dilute or even mask the presence of an aerosol source of interest. Furthermore, computational complexity limits the number of simulations which can be performed and reduced order modeling is difficult due to the advective nature of the problem. Noise and variability in the climate system is traditionally managed by temporal and spatial averaging. However, such averaging makes it more difficult to distinguish the source from background and hence is not viable in our context. We have addressed the challenges of noise and variability through a combination of techniques. Limited variability ensembles reduce the uncertainty due to internal atmospheric variability so that an accurate time evolution operator may be learned from limited data. By using source tagging, our simulations disentangle the aerosol source of interest from other background sources thus facilitating clean data to learn the aerosol plume evolution while capturing background aerosols statistics that may be incorporated in the inverse problem. Our use of localizing nonlinear dimension reduction techniques overcame the challenge of modeling advective phenomena thus making it possible to learn a reduced order model efficiently. To facilitate model reliability when used within an inversion framework, we designed an operator architecture that strictly enforces first principles chemistry such as conservation of molar mass and irreversibility of the $SO_2$ to sulfate reaction. We also developed chemistry-based validation metrics using reaction rate statistics over large sample sets to improve generalizably. Our Bayesian approximation error approach systematically accounts for both internal climate variably and background aerosol uncertainty to provide a reliable inversion framework which can accommodate observational data with unseen wind fields and background aerosols that are within the training data distribution. To the best of our knowledge, this combination of techniques is a first-of-its-kind approach to enable source inversion which is robust to both wind variability and background aerosol noise.

This article proposes a comprehensive framework to enable stratospheric aerosol source estimation with associated uncertainty. Although each aspect of our framework was designed based on the characteristics of stratospheric aerosol transport, there are potentially many other areas where it may be impactful. Our framework addresses challenges posed by internal variability and global spatial scales inherent in many problems arising from the earth sciences. Since we focused on AOD observational data, our approach is extensible to a variety of other chemical species which may be relevant in global atmospheric monitoring, climate attribution, or geoengineering.

Our results used an observational dataset synthesized from a test set simulation of the eruption of Mount Pinatubo. Since this was a large eruption, the signal due to the volcanic aerosols rose significantly above the background aerosols. For eruptions of smaller magnitude, our framework is applicable, but a longer time horizon may be needed to attain sufficient information from the AOD measurements. To extend the time horizon, it is necessary to take more RBF basis functions and thus introduce additional complexity in the RBF fitting and time evolution operator learning. Similarly, extending from 1D longitudinal data to 2D spatial data will require additional RBF basis functions which have more hyperparameters. Our framework is extensible and future research should explore finer RBF resolution and use in two spatial dimensions. To achieve this, the enforcement of additional constraints such as monotonicity of RBF hyperparameters (as a function of time) will be crucial to ensure identifiably.

Our uncertainty estimates are based on the Laplace approximation of the Bayesian posterior. Although this is pragmatic and common in practice, better uncertainty quantification is possible through the use of Markov Chain Monte Carlo methods. Future research should utilize the computational efficiency of the learned operators and availability of derivative information to enable more advanced sampling algorithms. This has the potential to realize full characterization of uncertainty rather than the Gaussian approximation used in this article.

\acknowledgements

The authors thank Hailong Wang and Yang Yang for sharing their source tagging code which provided a foundation for the implementation used in this article.

Data from the full E3SMv2-SPA simulation campaign including pre-industrial control, historical, and Mt. Pinatubo ensembles will be hosted at Sandia National Laboratories with location and download instructions announced on https://www.sandia.gov/cldera/e3sm-simulations-data/ when available.

This research used resources of the National Energy Research Scientific Computing Center (NERSC), a Department of Energy Office of Science User Facility using NERSC award BER-ERCAP0026535.

This work was supported by the Laboratory Directed Research and Development program at Sandia National Laboratories, a multimission laboratory managed and operated by National Technology and Engineering Solutions of Sandia LLC, a wholly owned subsidiary of Honeywell International Inc. for the U.S. Department of Energy’s National Nuclear Security Administration under contract DE-NA0003525. This written work is authored by employees of NTESS. The employees, not NTESS, own the right, title, and interest in and to the written work and is responsible for its contents. Any subjective views or opinions that might be expressed in the written work do not necessarily represent the views of the U.S. Government. The publisher acknowledges that the U.S. Government retains a non-exclusive, paid-up, irrevocable, world-wide license to publish or reproduce the published form of this written work or allow others to do so, for U.S. Government purposes. The DOE will provide public access to results of federally sponsored research in accordance with the DOE Public Access Plan. SAND2024-11582O.

 \appendix

\section{Selection of zonal wind rank} \label{appendix:wind_modes}

This section describes our analysis to optimize the number of PCA modes for the zonal wind dimension reduction. We preselected a set of candidate ranks $\{2, 4, 5,7\}$ based on significant jumps in the singular value magnitude seen in Figure~\ref{fig:zonal_wind_pca}.

\begin{figure}[h]
\centering
  \includegraphics[width=0.5\textwidth]{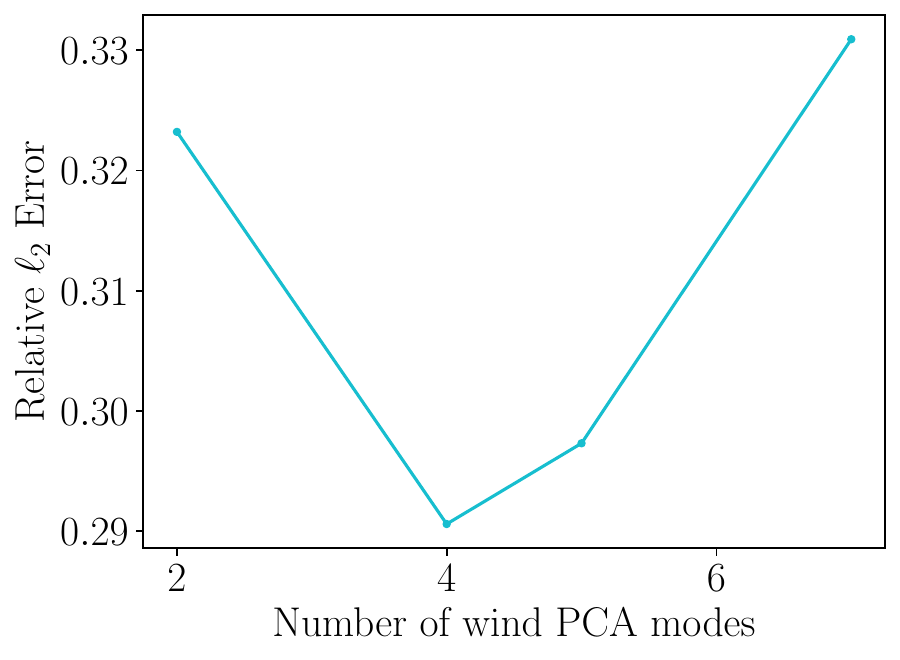}
    \caption{Network prediction error on the validation set as a function of the number of wind PCA modes. Each point is the average over 5 training instances with random instantiations of the network weights.}
  \label{fig:wind_pca_tune}
\end{figure}

We evaluate the number of PCA modes as a function of the time evolution operator prediction accuracy, measured using the relative $\ell_2$ prediction error in the for both $SO_2$ and sulfate over all timepoints over all 5 ensembles in the validation set.  We fixed the neural network to consist of a single linear layer as defined in~\eqref{eqn:NN_arch}.  The results are shown in Figure~\ref{fig:wind_pca_tune}. For each datapoint in the figure, we ran 5 instances of the neural network configuration with random initialization of the network weights, and display the average prediction error in the figure.  

The results match our intuition: there is a tradeoff between accurate representation of the wind (2 modes appears insufficient), and removing higher frequency wind variations that have minimal impact on the plume transport  (7 modes may not sufficiently filter the data). Based on the results, we truncated the reduced dimension zonal wind data to 4 modes. 

\section{Network architecture analysis} \label{appendix:depth_analysis}

This appendix considers analysis of the network architecture in terms of network depth, network width, and the learning rate schedule. A deeper network is more expressive and consequently deep networks have become common in many applications. However, the depth must be commensurate to the training data available as adding hidden layers results in additional model parameters which must be trained.  If the training data is insufficient for a given depth and width, the resulting network will likely perform well on the training set, but not generalize well. We note that similar concerns regarding overparameterization motivated our choice of fully connected networks over larger, potentially more expressive operator networks. 

 Evaluation based upon a small validation set, as in this article where the training and validation data comes from E3SM simulations, may be insufficient to detect poor generalization. Since we are training the network in the service of constraining an inverse problem, it is crucial that the network predictions remain physically plausible for samples outside of the training set. Our validation set cannot cover the full range of inputs relevant for the inverse problem. Hence we will explore metrics based on physical principles which can be computed for a larger set of input samples.

Since the inverse problem seeks to estimate the initial $SO_2$ using observations of the AOD evolution, the rate at which $SO_2$ transforms into sulfate is crucial to inform the source magnitude estimation. This led us to use the $SO_2$ depletion rate as a metric to assess model quality and guide our choice of the number of hidden layers. To measure the depletion rate from daily data, we consider the classical linear model for a chemical reaction
\begin{align} \label{eqn:linear_reaction_model}
\frac{d \alpha}{dt}(t) = \lambda \alpha(t),
\end{align}
where $\alpha(t)$ is the mass of the chemical species and $\lambda \in \R$ is the reaction rate. Since our data is at a daily time resolution, we approximate the time derivative with the difference of $\alpha$ evaluated at successive days. Letting $\alpha_n \in \R$ denote the total mass of $SO_2$ at day $t_n$, approximating $\frac{d \alpha}{dt}(t)=(\alpha_{n+1}-\alpha_n)/1$, and solving for the reaction rate at time $t_n$, we have
\begin{align} \label{eqn:reaction_rate}
\lambda_n = \frac{\alpha_{n+1}-\alpha_n}{\alpha_n}.
\end{align}
The linear reaction model~\eqref{eqn:linear_reaction_model} fails to fully capture the nonlinear evolution of $SO_2$. However,~\eqref{eqn:reaction_rate} provides a measure of the reaction rate locally at a given time step. Computing~\eqref{eqn:reaction_rate} at each time step gives an estimate of the reaction rate which, as demonstrated below, is sufficient to identify nonphysical behavior of models which generalize poorly \footnote{We note that the assumption of a fairly constant linear reaction rate is reasonable for the particular E3SM model used in this article that is without full chemistry and assumes infinite hydroxyl radicals.}. 

Computing the reaction rate at each time step (excluding the final since we cannot look ahead to estimate the time derivative) in the training set gives a baseline estimate of the range of reaction rates which are physically plausible.  To assess the quality of a trained network, we generate $1000$ initial $SO_2$ RBF coordinates and time series of zonal wind coordinates. The sampling distribution for the $SO_2$ RBF coordinates was defined by computing the range of initial $SO_2$ RBF coordinates over the $5$ validation set simulations (which had eruption magnitudes ranging from $3$ Tg to $15$ Tg) and defining a uniform distribution over an interval whose endpoints correspond to widening the validation set range by $10\%$ on both the minimum and maximum values. We sample the time series of zonal winds using a uniform distribution defined over the range of the training set zonal winds. We use the set of training zonal winds rather than the validation set because in the inverse problem, the training data zonal wind defined the samples used in the likelihood evaluation. For a given model, we generate $1000$ trajectories of $SO_2$ and compute the reaction rate~\eqref{eqn:reaction_rate} for each time step and each sample. This yields $1000$ time series of reaction rates based on data predicted by the network. We then compare the range of reaction rates in the training data (which represents a physically plausible baseline) with the range of reaction rates computed from the network predictions.

We considered networks with $0$, $1$, and $2$ hidden layers, and with widths $7$, $14$, and $21$ (for the $1$ and $2$ hidden layer networks). We included a batch normalization layer in the network with 2 hidden layers to improve network training.  For each depth-width pairing, we trained $6$ networks from different initializations and with various learning rate schedules.  For each network, we computed the standard deviation of the reaction rate time series for each sample, and averaged over the set of 1000 samples.  The lowest mean standard deviation of the reaction rates for each depth/width pairing is presented in Table~\ref{tab:reaction_rate}.

\begin{table}[!ht]
\centering
\begin{tabular}{|l|c|c|c|}
\hline
         & \multicolumn{1}{l|}{0 Hidden Layers} & \multicolumn{1}{l|}{1 Hidden Layer} & \multicolumn{1}{l|}{2 Hidden Layers} \\ \hline
Width 7  & \textbf{0.000716}            & 0.001007                     & 0.001312                     \\ \hline
Width 14 & -                            & 0.001107                     & 0.001309                     \\ \hline
Width 21 & -                            & 0.001055                     & 0.002252                     \\ \hline
\end{tabular}
\captionsetup{justification=centering}
\caption{Reaction rate standard deviation averaged over 1000 samples for fully connected neural networks with varying widths and number of hidden layers. }
\label{tab:reaction_rate}
\end{table}

\begin{figure}[h]
\centering
  \includegraphics[width=0.7\textwidth]{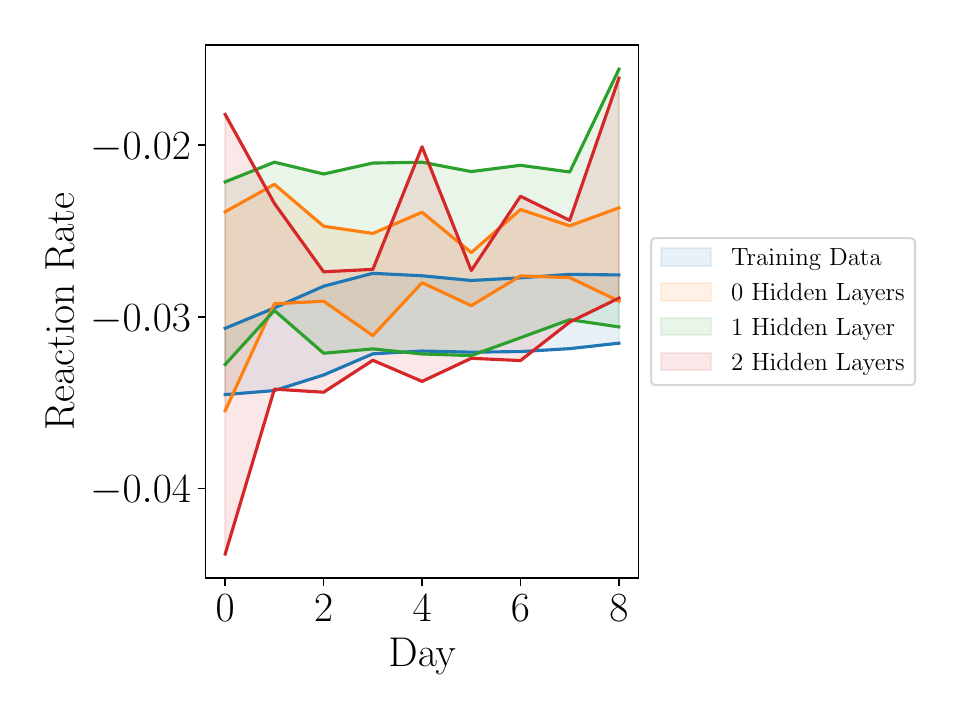}
    \caption{Spread of reaction rates in the time series corresponding to the training data and network predictions with varying network depths.}
  \label{fig:depth_analysis}
\end{figure}

We did not observe a notable trend on the impact of network width on the reaction rate. However, there was a trend in the depth affecting the reaction rate. Figure~\ref{fig:depth_analysis} displays the spread of the reaction rate time series for the training set, and the predicted reaction rates from the network with lowest mean standard deviation for the $0$, $1$, and $2$ hidden layer networks.  We summarize the spread of reaction rates over the sample set by shading the region bound by their minimum and maximum values at each time step. The $0$ hidden layer networks have a wider band compared to the training set and a slightly shifted mean reaction rate. The $1$ and $2$ hidden layer networks have much larger bands, particularly at the initial and final time steps. 

The analysis presented in this appendix led to our choice of a $0$ hidden layer network with the minimum reaction rate variability, as was presented in Section~\ref{sec:numerical_results}. We  note that while the network itself is linear, the full model is autoregressive with added nonlinear layers, as described in Section~\ref{sec:numerical_results}.

%% appendix sections are then done as normal sections and after Acknowledgements



%\begin{thebibliography}{-8}


%\small{
%\bibitem{key}

%...

%}

%\end{thebibliography}


\bibliographystyle{Bibliography_Style}

\bibliography{References}

\end{document}